%% file: acl_latex.tex
\pdfoutput=1
\documentclass[11pt]{article}

\usepackage{acl}

\usepackage{times}
\usepackage{latexsym}
\usepackage{graphicx}
\usepackage{booktabs}
\usepackage{multirow}
\usepackage{arydshln}
\usepackage{comment}
\usepackage{booktabs}

\usepackage{amsmath}
\usepackage{arydshln}
\usepackage{capt-of}% or \usepackage{caption}
\usepackage{booktabs}
\usepackage{varwidth}
\usepackage{multirow}
\usepackage{booktabs,arydshln}
\usepackage[skins]{tcolorbox}
\newtcolorbox{mybox}{enhanced,label=takeaways,colback=blue!5!white,colframe=red!75!black,size=title,bottom=0,title={\begin{center}\textbf{\large Summarized Takeaways}\end{center}},
boxed title style={enhanced,
  borderline={0.5mm}{-0.5mm}{red!75!black,solid},
  colframe=white,
  colback=red!75!black,
  colupper={black},
},}

\newtcolorbox{myframe}[2][]{%
 enhanced,colback=white,colframe=black,coltitle=black,boxrule=1pt,bottom=1pt, left=3pt,
  fonttitle=\itshape,
  attach boxed title to top left={yshift=-0.6\baselineskip-0.2pt,xshift=2mm},
  boxed title style={tile,size=minimal,left=0.5mm,right=0.5mm,
    colback=white,before upper=\strut},
  title=#2,#1
}

\makeatletter
\def\adl@drawiv#1#2#3{%
        \hskip.5\tabcolsep
        \xleaders#3{#2.5\@tempdimb #1{1}#2.5\@tempdimb}%
                #2\z@ plus1fil minus1fil\relax
        \hskip.5\tabcolsep}
\newcommand{\cdashlinelr}[1]{%
  \noalign{\vskip\aboverulesep
           \global\let\@dashdrawstore\adl@draw
           \global\let\adl@draw\adl@drawiv}
  \cdashline{#1}
  \noalign{\global\let\adl@draw\@dashdrawstore
           \vskip\belowrulesep}}
\makeatother

\usepackage{makecell}

\usepackage[T1]{fontenc}
\usepackage[utf8]{inputenc}

\usepackage{microtype}

\usepackage{CJKutf8}
\usepackage{todonotes}

\input{macros}

\DeclareMathOperator*{\argmax}{arg\,max}

\title{How to Handle Different Types of Out-of-Distribution Scenarios in Computational Argumentation? \\ A Comprehensive and Fine-Grained Field Study}
\author{Andreas Waldis\thanks{* Corresponding author andreas.waldis@live.com} $^{1,2}$, Yufang Hou$^{1,3}$, Iryna Gurevych$^{1}$ \\
$^1$Ubiquitous Knowledge Processing Lab (UKP Lab) \\
Department of Computer Science and Hessian Center for AI (hessian.AI)\\
Technical University of Darmstadt\\
 $^2$Information Systems Research Lab, Lucerne University of Applied Sciences and Arts \\
 $^3$IBM Research Europe - Ireland \\
\texttt{\href{http://www.ukp.tu-darmstadt.de/}{www.ukp.tu-darmstadt.de}} \hspace{0.5em} \texttt{\href{http://www.hslu.ch/}{www.hslu.ch}}\\
}
\begin{document}
\maketitle

\vspace*{0.25cm}
\begin{abstract}
The advent of pre-trained Language Models (LMs) has markedly advanced natural language processing, but their efficacy in out-of-distribution (OOD) scenarios remains a significant challenge \citep{hupkes2023taxonomy}.
The field of computational argumentation (CA), modeling human argumentation processes, is notably impacted by these challenges because complex annotation schemes and high annotation costs naturally lead to resources barely covering the multiplicity of available text sources and topics.
Due to this data scarcity, generalization to data from uncovered covariant distributions is a common challenge for CA tasks like stance detection or argument classification.
This work systematically assesses LMs' capabilities for such OOD scenarios.  
%for the first time.
While previous work targets specific OOD types like topic shifts \citep{stab-etal-2018-cross} or OOD uniformly \citep{yuan2023revisiting}, we address three prevalent OOD scenarios in CA: \textit{topic shift, domain shift,} and \textit{language shift}. 
Our findings challenge the general superiority of in-context learning (ICL) for OOD. 
We find that the efficacy of such learning paradigms varies with the type of OOD.
Specifically, while ICL excels for domain shifts with heavy label divergences between train and test data, prompt-based fine-tuning surpasses for shifts when semantic differences prevail, like topic shifts. 
Navigating the heterogeneity of OOD scenarios in CA, our work empirically underscores the potential of base-sized LMs to overcome these challenges.
\footnote{We provide data and code at \href{https://github.com/UKPLab/acl2024-ood-compuational-argumentation}{online}.}

\end{abstract}

\newcommand\ALBERT{ALBERT}
\newcommand\BERT{BERT}
\newcommand\BART{BART}
\newcommand\ELECTRA{ELECTRA}
\newcommand\DEBERTA{DeBERTa}
\newcommand\ROBERTA{RoBERTa}
\newcommand\GPT{GPT-2}
\newcommand\TfivethreeB{T5 (3B)}
\newcommand\FLANTfivethreeB{FLAN-T5 (3B)}
\newcommand\GPTNeo{GPT-Neo (2.7B)}
\newcommand\GLOVE{GloVe}

\renewcommand{\sectionautorefname}{§}
\renewcommand{\subsectionautorefname}{§}

\input{parts/1-Introduction.tex}

\input{parts/8-Related-Work.tex}

\input{parts/2-Methodology.tex}

\input{parts/2-Tasks.tex}

\input{parts/2-Prompt-based-Fine-Tuning.tex}
\input{parts/3-Experimental-Setup.tex}

\input{parts/4-Experiment.tex}

\input{parts/5-Analysis.tex}

\input{parts/7-Discussion.tex}

%\section*{Limitations}
\bibliography{anthology,custom}
\bibliographystyle{acl_natbib}

\input{parts/9-Appendix.tex}

\end{document}

%% file: macros.tex
\usepackage{todonotes}
\usepackage{booktabs}
\usepackage{comment}

\usepackage{amsmath}
\usepackage{amssymb}
\usepackage{arydshln}
\usepackage{booktabs}
\usepackage{tabularx}
\usepackage{tikz}
\usepackage{pgfplots}
\usepackage[normalem]{ulem}
\usepackage{xcolor}
\usepackage{dashbox}%
\usepackage{multirow}
\usepackage{textcomp}
\usepackage{tcolorbox}
\usepackage[ruled, vlined, linesnumbered]{algorithm2e}
\usepackage{mathtools}
\usepackage{soul}

\pgfplotsset{compat=1.13}

\definecolor{plot1}{RGB}{93,147,191}
\definecolor{plot2}{RGB}{233,72,73}
\definecolor{plot3}{RGB}{113,191,110}
\definecolor{plot4}{RGB}{163,73,151}
\definecolor{plot5}{RGB}{230,130,50}
\definecolor{decentgrey}{RGB}{232,232,232}

\usetikzlibrary{calc,fit,positioning,arrows}

\newtcbox{\pattern}{on line,colback=decentgrey,colframe=white,size=fbox,arc=3pt, box align=base, before upper=\strut, top=-2pt, bottom=-2pt, boxrule=0pt}

\pgfdeclarelayer{bg}    % declare background layer
\pgfsetlayers{bg,main}  % set the order of the layers (main is the standard layer)

%% file: parts/1-Introduction.tex
\begin{comment}
\todo{What are your main claims? - XYZ}
\todo{Why is the problem important? – It is important because}
\todo{Why is it difficult? – it is difficult because}
\todo{How others approach it and where they fail? -> not considering topic, domain, language at the same time}
\todo{What are the novel findings? What do we know now what we did not know before? – hard for me to answer, what are you the first one to discover?}
\todo{How do you explain your findings?}
 
\end{comment}

\section{Introduction}

\begin{figure}[t]
 \centering
\vspace*{0.5cm}
 \includegraphics[width=0.48\textwidth]{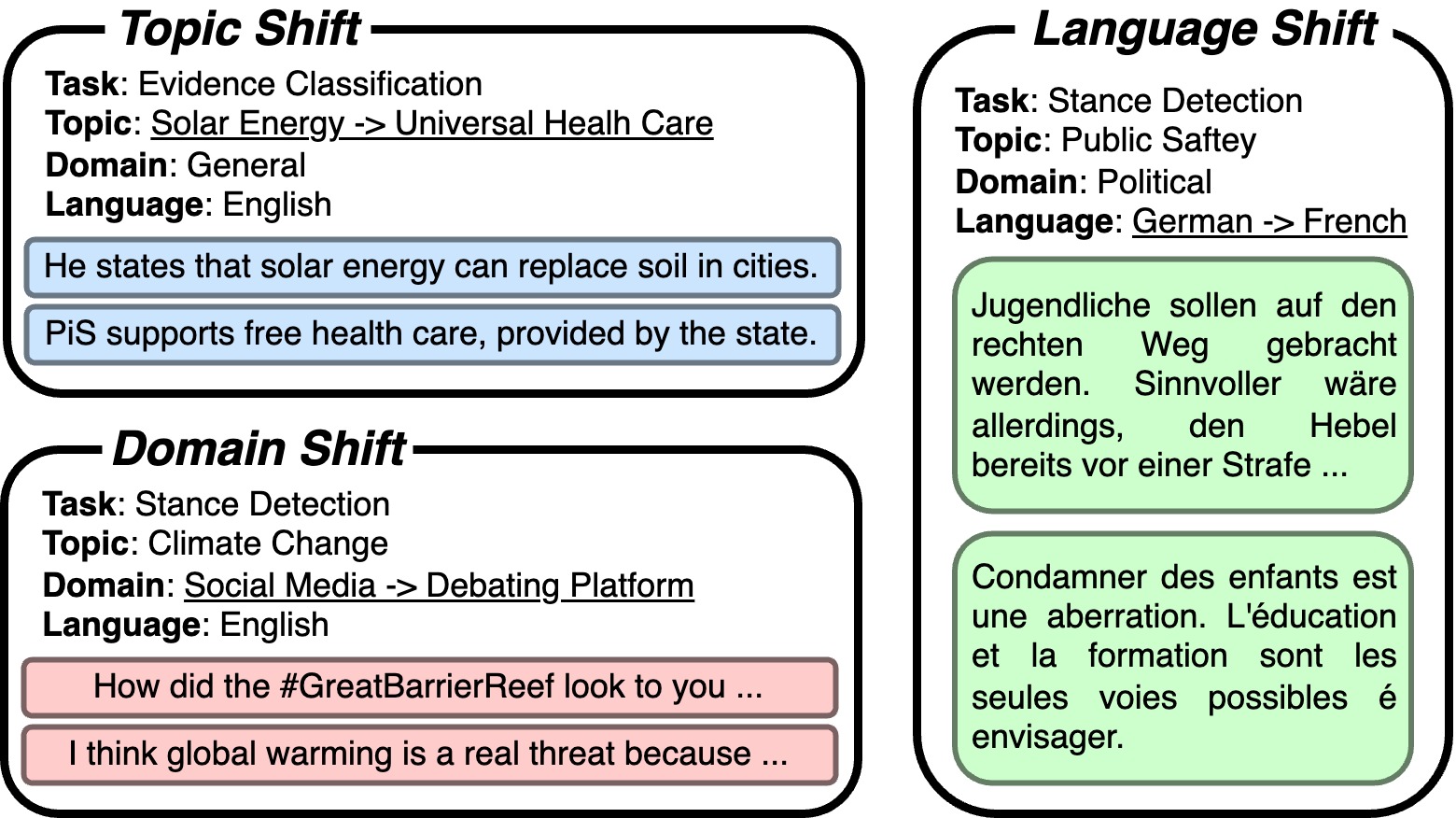}
 \caption{Common OOD types of computational argumentation covering \textit{evidence classification} \citep{shnarch-etal-2018-will}, and mono or multilingual \textit{stance detection} \citep{Hardalov2021CrossDomainLS,DBLP:conf/swisstext/VamvasS20}.}
 \label{fig:shifts}
\end{figure}

Argumentation as a communication tool for human reasoning has engaged researchers over millennia \citep{Aristotle2006-ARIORA,Toulmin1960TheUO,van2019handbook} and has become an important research area in natural language processing under the umbrella of \textit{computational argumentation} \citep{lippi2016argumentation,lauscher2022scientia}.
Specifically, computational argumentation (CA) models human argumentative processes and leads to complex tasks such as stance detection \cite{mohammad-etal-2016-semeval} and argument quality evaluation \cite{toledo-etal-2019-automatic}.
However, developing resources for such CA tasks requires significant annotation efforts \citep{10.1162/COLI_a_00276,schiller2022effect}, which often inadequately capture the wide range of heterogeneity in available text sources and topics.
This situation makes OOD scenarios, especially those involving significant \emph{covariant distribution shifts}, a common challenge for CA tasks since LMs are anticipated to generalize across such shifts in current and future applications \citep{slonim2021autonomous}.
These shifts occur when input data distribution changes between the training and testing phrases and can be viewed as a specific aspect of out-of-distribution (OOD) scenarios \citep{Zhang2020DiveID}. 

This work focuses on three types of covariant distribution shifts frequently encountered in CA tasks: \emph{topic shift, domain shift}, and \emph{language shift}. 
\autoref{fig:shifts} illustrates these three types of OOD scenarios, in which researchers aimed at developing systems for CA tasks that generalize across unseen topics \citep{shnarch-etal-2018-will,toledo-etal-2019-automatic}, text domains \citep{lauscher-etal-2020-rhetoric,Hardalov2021CrossDomainLS}, or languages \citep{eger-etal-2018-cross,DBLP:conf/swisstext/VamvasS20}. 
These studies have observed that CA systems often fail to handle OOD scenarios.

Given this variety of OOD scenarios in CA and the need for data efficiency, this paper aims to answer the following research question: ``\emph{how to handle different types of OOD scenarios in computational argumentation using LMs}''? 
Most previous work on evaluating the generalization and robustness of NLP models has either predominantly focused on a single type of OOD scenario, such as domain shift \cite{DBLP:conf/acl/BlitzerDP07,Hardalov2021CrossDomainLS}, or on general OOD that does not distinguish among various types of OOD scenarios \citep{yuan2023revisiting}. 
However, these studies overlook the heterogeneous nature of OOD and, thereby, limit the transferability of the corresponding findings to the spectrum of shift types in CA tasks.  
This study introduces a detailed evaluation framework encompassing holistic performance measures (\autoref{sec:methodology}) to pinpoint crucial generalization flaws such as misalignment between performance and training loss in models. 
In addition, we feature a heterogeneous collection of eleven CA tasks (\autoref{sec:ood-tasks}) covering three types of OOD scenarios. 
We evaluate these tasks with an extensive experimental setup (\autoref{sec:experiments}) covering twelve LMs of various sizes and eight learning paradigms, including gradient-based learning like vanilla fine-tuning (FT) and prompt-based fine-tuning (P+FT), as well as in-context learning (ICL).
From the observed results (\autoref{sec:results}), we conduct an in-depth analysis (\autoref{sec:analysis}) to understand better how learning paradigms and LMs differ for different types of OOD for computational argumentation. 

In contrast to \citet{yuan2023revisiting}, suggesting in-context learning (ICL) surpasses fine-tuning LMs for addressing general OOD, we find different learning paradigms excel in different types of OOD for CA tasks. 
In particular, ICL outperforms gradient-based learning for domain shifts where train and test label distributions heavily differ.
However, gradient-based learning surpasses ICL for topic shifts characterized by a clear semantic divergence in the covered topics between the training and testing datasets.  %with clear semantic divergences between train and test data as distinct topics are covered.

\paragraph{Contribution}
We summarize our work regarding four contributions:
\begin{enumerate}
    \item \textbf{Evaluation} We propose an evaluation framework including eleven CA tasks across three types of OOD scenarios.
    Along with a comprehensive assessment of LM's OOD capabilities, it provides a clear picture of the generalization challenges in CA and %guides
    offers guidance to 
    practitioners in tackling these challenges.
    
    \item \textbf{Results} Extensive experiments offer valuable insights and show that different learning paradigms effectively manage OOD scenarios for CA under different conditions. Particularly, 
    in-context learning should be preferred for domain shifts, while gradient-based learning is the first choice for %generalizing
    generalization 
    across semantic differences (topic shifts). 
    
    \item \textbf{Analysis} We shed light on the unused potential of base-sized LMs for OOD scenarios.
    We demonstrate that training a fraction of the parameters of base-sized LMs with LoRA achieves performance comparable to full LM tuning, and such parameter-efficient training offers better stability than larger LMs.

    \item \textbf{Facilitating Research} This work emphasizes the critical role of OOD heterogeneity in tackling generalization challenges within CA tasks.
    This paves the way for future research to conduct detailed and targeted examinations of OOD scenarios in other research areas.
\end{enumerate}

%% file: parts/8-Related-Work.tex
\section{Related Work}
\label{sec:related_work}

\paragraph{Out-of-Distribution Generalization}
Studies in NLP target OOD generalization from different perspectives, focusing on the robustness of LMs \citep{DBLP:conf/icml/HendrycksLM19, DBLP:conf/aaai/JinJZS20, DBLP:conf/nips/ZhouXGM0W20,DBLP:conf/iclr/WangWCGJLL21} or OOD detection \citep{DBLP:conf/bmvc/KonerSRGT21, Cho2023ProbingOR}.
Similar to \textit{computer vision} \citep{DBLP:conf/iclr/TsengLH020}, NLP studies primarily focus on considering covariant distribution shifts \citep{Zhang2020DiveID} and analyze single types of them in isolation, such as domain across datasets \citep{Hardalov2021CrossDomainLS,yang-etal-2023-glue,yuan2023revisiting}, language \citep{DBLP:conf/iclr/KWMR20, DBLP:conf/acl/ConneauKGCWGGOZ20}, topic \citep{stab-etal-2018-cross,allaway-mckeown-2020-zero}.
This shortage of comprehensively analyzing OOD hinders analytical or methodological advancements in a challenging field such as \textit{computational argumentation} since generalization of methods is limited when relying on shift-specific features \citep{liang-etal-2022-jointcl,xu-etal-2018-cross,peng-etal-2018-cross,DBLP:conf/lrec/RietzlerSOE20}.

\paragraph{Prompt-based Fine-tuning}
Commonly, pre-trained LMs are fine-tuned by providing a natural language input and optimizing regarding an arbitrary label \citep{devlin-etal-2019-bert}.
Instead, prompt-based fine-tuning \cite{DBLP:journals/corr/abs-2107-13586} (P+FT) allows relying upon acquired competencies during pre-training, both for encoding the input and predicting the label by formulating the task as a cloze test.  
This procedure allows LMs to reach comparable performance to their large-sized counterparts \citep{schick-schutze-2021-exploiting,schick-schutze-2021-just} with the same limited data as in few-shot settings.
Despite their success for few-shot scenarios, little work analyzed how P+FT generalizes differently than FT or how it performs considering complete datasets, exceptionally \citet{DBLP:journals/corr/abs-2303-07320} showed the robustness of P+FT against adversarial attacks. 

\begin{figure}[t]
 \centering
 \includegraphics[width=0.48\textwidth]{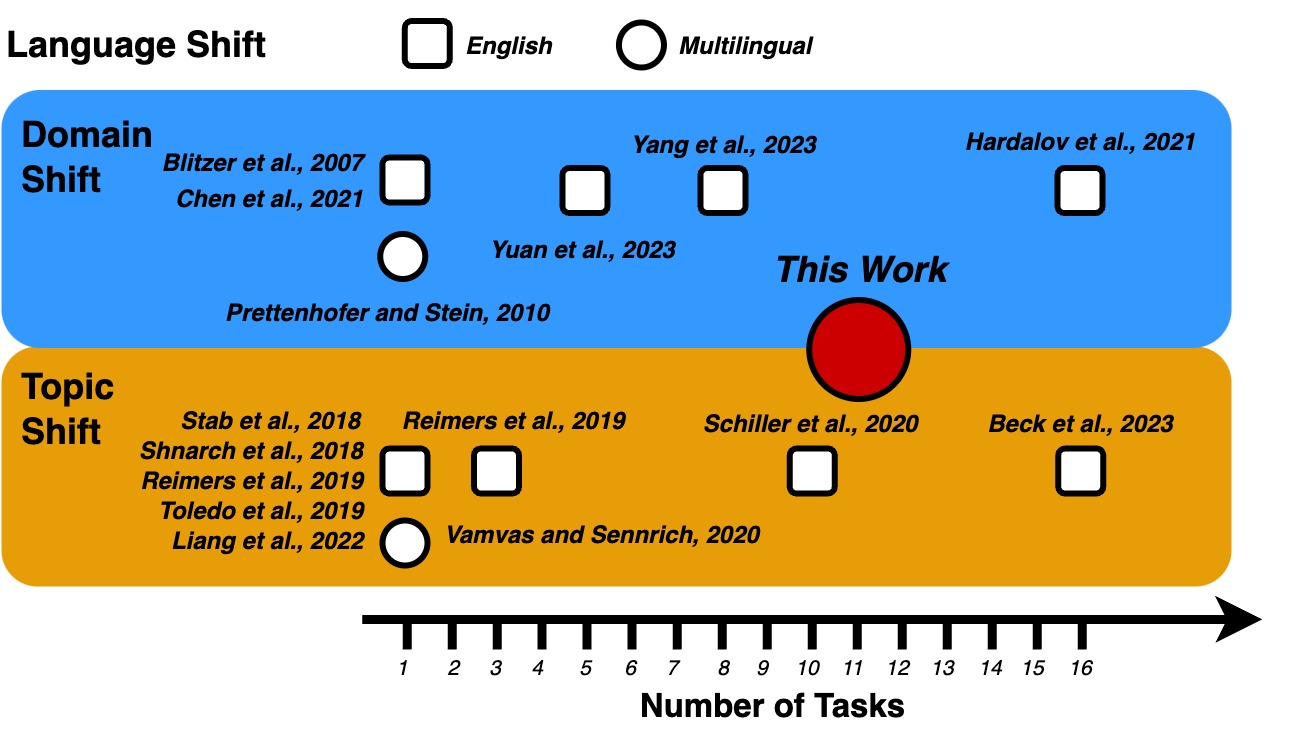}
 \caption{Comparison of our study, covering topic (orange), domain (blue), or language (square/circle) covariant distributions shifts with previous studies that mainly consider single shifts.}%The x-axis indicates the number of considered tasks.}
 \label{fig:covered_tasks}
\end{figure}

With this work (\autoref{fig:covered_tasks}), we address the need to comprehensively evaluate OOD abilities of LMs with a particular focus on \textit{computational argumentation}.
In particular, we assess a variety of LMs using in-context and gradient-based learning paradigms, considering %their
three 
types of OOD scenarios covering eleven different tasks.

%% file: parts/2-Methodology.tex
\section{Methodology}
\label{sec:methodology}

\subsection{OOD Types}\label{subsec:generalization-scenarios}
We distinguish between two generalization scenarios: in-distribution (ID) and out-of-distribution (OOD)   generalization.
While ID assumes train and test data being independent and identically distributed, OOD accounts for practical challenges where we expect apparent distribution shifts between the training and testing instances \citep{shen2021towards}.
To capture the success of a classifier $f(y|x)$ in such scenarios, we measure its ability to transfer learning from train instances $X_{train}$ to test instances $X_{test}$.
However, OOD potentially introduces covariant, label, and concept shifts between train and test data \citep{Zhang2020DiveID}.
In this work, we focus on three types of covariant shifts (\emph{topic shift, domain shift}, and \emph{language shift}), as illustrated in \autoref{fig:shifts}, due to their frequent prevalence in computational argumentation tasks. 

\subsection{OOD Evaluation Protocol}\label{subsec:evaluation}
Generalization success is typically measured with single task metrics, like the $F_1$ macro score. 
However, solely relying on one metric ignores known stability issues, such as apparent deviations regarding randomness \citep{DBLP:conf/iclr/MosbachAK21}. 
Thus, we compose a set of three requirements that a superior learning model should fulfill: good task performance (\textit{Applicability}), better alignment between optimization and evaluation (\textit{Reliability}), and \textit{Stability} regarding data and randomness.
We ground this evaluation for a specific task on a given set of runs ($r \in R$), trained for one distinct fold and seed over a number of epochs ($e \in E$).
Note that we formalize these requirements with OOD classification in mind and, therefore, rely on $F_1$ macro score as the reference metric.
However, these requirements can be generalized to other types of OOD tasks using the corresponding reference metrics, such as ROUGE for text generation.

\paragraph{Applicability} captures the task-specific performance. 
Specifically, we measure the average task-specific metric, here $F_1$ macro score ($\mu_{F_1}$), across all runs $r$ covering different folds and seeds. 

\paragraph{Reliability} requires that the \textit{learning} process (optimization objective) is reflected in the obtained task \textit{performance}.
We evaluate the model using the development dataset that embodies the same OOD type as the training dataset. 
Specifically, we approximate, after each epoch $e$ of a run $r$, \textit{learning} as the loss $\beta=\{\forall e \in E(r) | f_{\text{loss}}^{\text{dev}}(e)\}$ and \textit{performance} using task metric ($F_1$ macro) $\gamma=\{\forall e \in E(r) | f_{F_1}^{\text{dev}}(e)\}$. 
Then, we calculate the Kendall correlation between $\beta$ and $\gamma$ and average it for every $r$ as $\mu_{\tau}$.
Ideally, we expect a negative correlation ($\tau=-1$), indicating that improvements in \textit{learning} are reflected in better \textit{performance} and vice-versa.
However, since we determine final labels using the argmax operation, dev loss and performance can increase simultaneously. 
For example, while predicting the same class ($\hat{y} = c_0$) the class probabilities can change from $(95\%, 5\%)$ to $(90\%, 10\%)$. 
At the same time, the cross-entropy changes from $0.074$ to $0.15$.
Therefore, we assume the model is becoming less sure about the prediction.
This aspect is particularly relevant for OOD generalization, where overfitting to distributional properties of training data, such as unique vocabulary, likely introduces uncertainty during inference.

\paragraph{Stability} demands a low impact from varying data and randomness on both \textit{Applicability} and \textit{Reliability}.
As recommended by \citet{reimers-gurevych-2017-reporting}, we measure the standard deviation of $\sigma_{F_1}$ and $\sigma_{\tau}$ across $R$ runs covering different data folds and seeds.

%% file: parts/2-Tasks.tex
\section{CA Tasks Across OOD Types}\label{sec:ood-tasks}
In this section, we present the selection of computational argumentation tasks (\autoref{subsec:task-selection}) and subsequently show their heterogeneous distribution shifts, focusing on covariant and label properties (\autoref{subsec:task-characteristic}).

\subsection{Task Selection}\label{subsec:task-selection}
We choose eleven tasks from computational argumentation and related fields \citep{stede2020automatic} that inherent OOD as a fundamental challenge.
We broadly categorize them according to their targeted covariant distribution shift, either \textbf{topic}, \textbf{domain}, or \textbf{language}.
For example, \textit{domain} for stance detection across datasets \citep{Hardalov2021CrossDomainLS}, sentiment analysis across \textit{languages} \citep{prettenhofer-stein-2010-cross}, or argument quality across \textit{topics} \citet{toledo-etal-2019-automatic}. 
\autoref{fig:covered_tasks} compares our study with previous research in terms of the number of tasks  and the range of OOD types covered. 
Below we briefly describe each task: 
\paragraph{Argument Quality (\textit{arg-qua})} \citet{toledo-etal-2019-automatic} analyzed  9,100 argument pairs across \textbf{22 topics} to determine which one has higher quality.

\paragraph{Argument Similarity (\textit{arg-sim})} \citet{reimers-etal-2019-classification} annotated 3,595 arguments pairs of \textbf{28 topics} to decide whether they are similar or not. 

\paragraph{Argument Classification (\textit{arg-cls})} \citet{stab-etal-2018-cross} annotated the stance of arguments (\textit{pro}, \textit{con}, \textit{neutral}) regarding one of \textbf{eight topics}. 

\paragraph{Evidence Classification (\textit{evi-cls})} \citet{shnarch-etal-2018-will} presented 5,785 sentences annotated as relevant or not for one out of \textbf{118 topics}. 

\paragraph{Sentiment Classification (\textit{review})} \citet{DBLP:conf/acl/BlitzerDP07} annotated 8,000 reviews as positive or negative for \textbf{four domains} (Amazon product groups). 

\paragraph{Multi-Dataset Stance Detection (\textit{stance})} Following \citet{Hardalov2021CrossDomainLS}, we use the \textit{semeval} \citep{mohammad-etal-2016-semeval}, \textit{emergent} \citep{ferreira-vlachos-2016-emergent}, and \textit{iac} dataset \citep{DBLP:conf/lrec/WalkerTAAK12} to evaluate stance detection across \textbf{three domains} (\emph{social media, news}, and \emph{debating}). All of them provide the same labels (\textit{pro}, \textit{con}, \textit{neutral}). 

\paragraph{Multi-Dataset Entailment (\textit{entail})} Following \citet{yang-etal-2023-glue}, we consider three medium-sized datasets (\textit{rte} \citep{wang-etal-2018-glue}, \textit{SciTail} \citep{DBLP:conf/aaai/KhotSC18}, \textit{hans} \citep{mccoy-etal-2019-right}) to evaluate textual-entailment across \textbf{three domains}. 

\paragraph{Multi-Lingual Stance Detection (\textit{x-stance})} This dataset \citep{DBLP:conf/swisstext/VamvasS20} includes 63,000 multilingual comments (\textit{de}, \textit{fr}, \textit{it}) annotated as \textit{favor} or \textit{against} regarding \textbf{12 topics}.

\paragraph{Multi-Lingual Sentiment Classification (\textit{x-review})} \citet{prettenhofer-stein-2010-cross} presents a set of 43,000 positive or negative reviews covering \textbf{four languages} (\textit{de}, \textit{en}, \textit{fr}, \textit{jp}) and \textbf{three domains} (Amazon product groups). 

\vspace{5pt}

While the first seven English-only datasets mentioned above, include annotations for one considered shift (\emph{topic} or \emph{domain}), the selected multilingual datasets come with multiple such annotations.
This enables formulating four OOD tasks from two datasets addressing two shift types each: language and domain shifts for \textit{x-review} and topic and language shifts for \textit{x-stance}.

\subsection{Tasks Characteristics}\label{subsec:task-characteristic}
In this subsection, we delve into the nature of the distribution shifts embodied by the selected tasks.

\begin{table}[t]
\centering
    \setlength{\tabcolsep}{2pt}
    \resizebox{0.48\textwidth}{!}{%
        \begin{tabular}{l|c|c|cc|c}
        \toprule
            % & & & \bf \multicolumn{2}{c}{Overlap} & \bf \multicolumn{3}{c}{Average Train-Test Differences}\\
              &\bf Shift Type & \bf Separability  & \bf $\Delta$ Flesch  & \bf $\Delta$ Words & \bf KL\\
        \midrule
    \textit{arg-qua} & Top. & 78.6 & 1.5 & 2.2 & 0.1\\
    \textit{arg-sim} & Top. & 75.8 & 4.6 & 0.27 & 0.4\\
    \textit{arg-cls} & Top. & 28.7 & 2.0 & 0.6 & 1.6\\
    \textit{evi-cls} & Top. & 56.3 & 2.4 & 0.7 & 7.1\\ \cdashlinelr{1-6}
    \textit{review} & Dom. & 52.7 & 6.5 & 60.5 & 0.0\\
    \textit{stance} & Dom. & 86.7 & 2.7 & 60.8 & 70.8\\
    \textit{entail} & Dom. & 40.4 & 5.1 & 31.2 & 12.8\\ \cdashlinelr{1-6}
    \textit{x-stance} & Lang./Top. & 0.05/19.8 & 16.6/1.3 & 6.6/0.3 & 0.6/0.4\\
    \textit{x-review} & Lang./Dom. & 0.07/72.4 & 11.0/1.8 & 60.0/6.5& 0.0/0.0\\
        \bottomrule
        \end{tabular}
    }
    \caption{
    Distribution shift characteristics between train and test splits of the eleven tasks (averaged across all folds): 
    Separability, differences between train and test instances regarding Flesch score, number of words, and the class distribution (KL divergence).
    }
    \label{tab:task-details}
\end{table}

\paragraph{Shift Characteristics}
We focus on covariant properties of the input $x$ (such as semantics) and the label $y$ to describe the characteristic of distribution shifts between training and testing instances.
\autoref{tab:task-details} show these properties, with higher values denoting increased challenge levels. 
First, we assess the separability of train and test instances based on their semantic representation. 
Following \citet{DBLP:conf/icml/SunM0L22}, we embed\footnote{Following \citet{reimers-gurevych-2019-sentence}, we use \texttt{paraphrase-multilingual--mpnet-base--v2} for embedding.} all instances and apply k-means clustering \citep{DBLP:journals/tit/Lloyd82,macqueen1967classification} to form two clusters. 
The alignment of these clusters with the train/test split is measured using the adjusted Rand index \citep{hubert1985comparing}. 
A higher score suggests a more pronounced semantic shift between train and test sets. 
Subsequently, we examine biases in surface-level text features introduced during training by calculating differences in average readability \citep{flesch1948new} and word count ($\Delta$ Flesch, $\Delta$ Words) between training and testing instances. 
Furthermore, we evaluate distributional disparities in class labels using Kullback-Leibler (KL) divergence \citep{boyd2004convex}. 
Higher KL values indicate more pronounced imbalances, complicating the task, as LMs often develop biases towards the training label distribution.

\paragraph{Task Difficulties}
Drawing from the above analyses, we categorize tasks into distinct groups. 
\textit{arg-qua}, \textit{arg-sim}, and \textit{stance} demonstrate high semantic differences, with separability scores ranging between $75.8$ and $86.7$. 
Tasks like \textit{review}, \textit{stance}, \textit{entail}, and \textit{x-review} present surface-level challenges due to varying readability ($\Delta$ Flesch) and text lengths ($\Delta$ Word Count). 
Additionally, tasks such as \textit{evi-cls}, \textit{stance}, and \textit{entail} show notable label distribution imbalances, reflected in high KL divergence values, thereby adding further complexity.
Notably, \textit{stance} emerges as particularly challenging, exhibiting distinct semantic, surface form, and label differences between training and testing instances, coupled with significant divergence from the LMs' pre-trained text understanding.

%% file: parts/3-Experimental-Setup.tex
\section{Experimental Setup}
\label{sec:experiments}
This section outlines the experimental setup covering the models, learning paradigms, and evaluation.

\paragraph{Models}\label{sec:models}
We primarily experiment with base-sized LMs, including \textbf{BERT} \citep{devlin-etal-2019-bert}, \textbf{RoBERTa} \citep{Liu2019RoBERTaAR}, and \textbf{DeBERTa-v3} \citep{he2021deberta} and their multilingual counterparts \citep{devlin-etal-2019-bert,conneau-etal-2020-unsupervised,he2021deberta}.
For additional experiments, we consider base-sized version of \textbf{ALBERT} \citep{Lan2020ALBERTAL}, \textbf{DeBERTa} \citep{DBLP:journals/corr/abs-2111-09543}, \textbf{ELECTRA} \citep{DBLP:conf/iclr/ClarkLLM20}, and 3B version of \textbf{T5} \citep{raffel2020exploring} and  \textbf{FLAN-T5} \citep{chung2022scaling}.
Further we consider \textbf{GPT-3.5} \citep{DBLP:conf/nips/Ouyang0JAWMZASR22}, \textbf{Llama-2-Chat} (70B) \citep{touvron2023llama}, and \textbf{Orca-2} (13B) \citep{mitra2023orca}.
%, \textit{T5 (3B)}\citep{10.5555/3455716.3455856}, \textit{TK-Instruct (3B)} \citep{wang-etal-2022-super}, and \textit{PYTHIA (2.8B)}.

%\paragraph{Large Language Models (LLMs)}

\paragraph{Learning Paradigms}

We assess the generalization capabilities of LMs under various learning paradigms. 
This includes vanilla fine-tuning (\textbf{FT}), prompt-based fine-tuning (\textbf{P+FT}), and in-context learning (\textbf{ICL}).
Further, we consider linear probing (\textbf{LP}) and cloze prompting (\textbf{P}) as lower bounds to capture the LM's pre-trained capabilities.
In LP and FT, we train task-specific classification heads, in which the LM  remains either frozen (LP) or trainable (FT)\footnote{\label{note:sep}We use [SEP] to concatenate the input with its topic, if available}. 
For P and P+FT, we embed the input into a cloze and let the pre-trained MLM head to predict the masking token and keep the LM frozen (P) or trainable (P+FT).
In addition, we verify scaling gradient-based methods to bigger LMs using parameter efficient methods, including \textbf{LoRA} \citep{DBLP:conf/iclr/HuSWALWWC22}, \textbf{P-Tuning} \citep{DBLP:journals/corr/abs-2103-10385}, and \textbf{Prompt-Tuning} \citep{DBLP:conf/emnlp/LesterAC21}.
Finally, using ICL, we verify the capabilities of large LMs with task-specific instructions and demonstrations.
Appendix \autoref{subsec:prompt-based-fine-tuning} and \autoref{subsec:icl-setup} provide more details about these learning paradigms.
%Further, we examine in Appendix \autoref{subsec:lpft} an additional learning paradigm that pre-initializes the classification head using a linear probe \citep{Kumar2022FineTuningCD} but we find it did not exhibit improvements for OOD scenarios.

\paragraph{Evaluation}\label{subsec:generalization-properties}

We enforce distribution shifts for OOD evaluation by composing train/dev/test splits, including instances with distinct distributional properties, such as unique topics or text domains (\autoref{fig:shifts}).
We utilize multi-fold cross-validation (CV) to account for data variability and ensure each distinct distributional property (like a unique topic) is tested precisely once \footnote{See Appendix \autoref{subsec:appendix:folds} for more details.}.
We evaluate all tasks on all learning paradigms, taking LP and P as a lower bound and ID fine-tuning (FT-ID) as an upper bound.
We assess every task using three random seeds to account for randomness. 
Using these runs, we employ comprehensive performance measurement including average \textit{Stability} ($\mu_{F_1}$), \textit{Reliability} ($\mu_{\tau}$), and the \textit{Stability} ($\sigma_{F_1}$,$\sigma_{\tau}$) - as previously defined in \autoref{subsec:evaluation}.

%% file: parts/4-Experiment.tex
\begin{table*}[t]
\centering
  \setlength{\tabcolsep}{7pt}
  \resizebox{1.00\textwidth}{!}{%
    \begin{tabular}{l|cccc|ccc|cc|c|c}
    \toprule
       & \bf arg-qua & \bf arg-sim & \bf arg-cls & \bf evi-cls & \bf review & \bf stance & \bf entail & \bf x-stance & \bf x-review & \bf $\uparrow$ \textit{Applicability} & \bf $\downarrow$ \textit{Reliability}\\
     &  \it \small{Top.}  &  \it  \small{Top.} &  \it    \small{Top.}  &  \it  \small{Top.}   &  \it \small{Dom.}   &  \it  \small{Dom.}  &  \it   \small{Dom.}   &   \it    \small{Lang./Top.} &  \it     \small{Lang./Dom.} & $\mu_{F_1}\pm\sigma_{F1}$   &  $\mu_{\tau}\pm\sigma_{\tau}$    \\
    \midrule
 \textbf{LP}\textsubscript{BERT} &  48.4 &  57.1 &    42.7 &   65.6 &  81.0 &  27.9 &    46.3 & 52.5/56.7 &  67.5/73.3 & $56.3\pm\small{0.8}$ &  $-58.4\pm\small{6.2}$ \\
 \textbf{P}\textsubscript{BERT} &  40.5 &  50.4 &    40.1 &   49.2 &  72.9 &  25.0 &    41.2 & 34.5/48.6 &  45.6/54.5 & $45.7\pm\small{0.2}$ &  - \\
 \textbf{FT}\textsubscript{BERT} &  75.5 &  \underline{68.4} &    57.5 &   74.7 &  \underline{89.3} &  \underline{31.1} &    \underline{50.7} & \underline{62.0}/\underline{63.9} &  77.7/\underline{84.4} & $66.8\pm\small{0.9}$ & $-56.8\pm\small{12.3}$ \\
 \textbf{P+FT}\textsubscript{BERT} &  \underline{76.2} &  66.0 &    \underline{59.8} &   \underline{75.7} &  \underline{89.3} &  28.5 &    48.0 & 59.5/63.6 &  \underline{79.6}/83.9 & $66.4\pm\small{1.1}$ & $-61.7\pm\small{12.4}$ \\
\cdashlinelr{1-12}
    \textbf{FT-ID}\textsubscript{BERT}   &87.9 &  76.4 &    67.3 &   78.9 &  90.4 &  61.1 &    93.6 &   67.6 &     87.0 & $78.9\pm\small{0.4}$ & $-96.1\pm\small{6.5}$ \\ \midrule
\textbf{LP}\textsubscript{DeBERTa-v3} &  53.0 &  70.0 &    55.1 &   67.9 &  88.6 &  23.4 &    58.0 & 55.4/59.7 &  78.7/83.6 & $63.0\pm\small{0.5}$ &  $-64.3\pm\small{4.3}$ \\
\textbf{P}\textsubscript{DeBERTa-v3} &  54.2 &  58.6 &    40.3 &   57.2 &  61.9 &  26.5 &    54.6 & 51.1/51.2 &  49.5/52.0 & $50.6\pm\small{1.0}$ &  - \\
\textbf{FT}\textsubscript{DeBERTa-v3} &  78.4 &  75.4 &    64.0 &   77.3 &  93.4 &  29.6 &    55.6 &  \textbf{\underline{69.8}}/69.3 &  91.3/90.9 & $72.3\pm\small{1.1}$ & $-72.6\pm\small{13.4}$ \\
\textbf{P+FT}\textsubscript{DeBERTa-v3} &  \textbf{ \underline{78.5}} &  \textbf{\underline{79.1$\dagger$}} &     \textbf{\underline{74.6}} &   \textbf{\underline{78.6}} &   \textbf{\underline{94.2$\dagger$}} &   \textbf{\underline{33.0}} &     \textbf{\underline{60.2}} & 69.7/\textbf{\underline{69.9}} &  \textbf{\underline{91.8}}/ \textbf{\underline{91.4}} & $74.6\pm\small{0.9}$ &  $-78.4\pm\small{8.4}$ \\ \cdashlinelr{1-12}
    \textbf{FT-ID}\textsubscript{DeBERTa-v3}   & 89.0 &  78.4 &    75.2 &   80.6 &  93.9 &  63.3 &    95.4 &   72.2 &     92.1 & $82.2\pm\small{0.4}$ & $-97.7\pm\small{6.5}$ \\ \midrule
   \textbf{LP}\textsubscript{RoBERTa} &  51.8 &  55.3 &    41.6 &   62.5 &  85.7 &  28.7 &    39.2 & 55.1/57.5 &  82.8/82.5 & $58.4\pm\small{0.6}$ &  $-56.3\pm\small{6.2}$ \\
   \textbf{P}\textsubscript{RoBERTa}&  48.3 &  55.3 &    42.9 &   51.8 &  80.5 &  24.0 &    40.9 & 42.4/48.7 &  67.2/73.4 & $52.3\pm\small{0.0}$ &  - \\
   \textbf{FT}\textsubscript{RoBERTa}&  70.9 &  73.0 &    56.9 &   77.5 &  \underline{92.2} &  \underline{30.0} &    51.3 & 62.2/66.8 &  89.6/\underline{90.1} & $69.1\pm\small{2.5}$ & $-69.7\pm\small{10.4}$ \\
   \textbf{P+FT}\textsubscript{RoBERTa} &  \underline{77.6} &  \underline{74.3} &    \underline{66.0} &   \underline{77.9} &  92.0 &  29.1 &    \underline{52.4} & \underline{67.4$\dagger$}/\underline{67.5$\dagger$} &  \underline{89.7}/90.0 & $71.3\pm\small{0.5}$ &  $-75.5\pm\small{8.1}$ \\\cdashlinelr{1-12}
    \textbf{FT-ID}\textsubscript{RoBERTa}   & 84.0 &  79.4 &    71.0 &   80.9 &  92.9 &  64.7 &    94.1 &   66.3 &     91.0 & $80.5\pm\small{1.9}$ & $-96.6\pm\small{4.7}$ \\ 
  \bottomrule
    \end{tabular}
  }
  \caption{OOD results using linear probing (\textbf{LP}), prompting (\textbf{P}), vanilla fine-tuning (\textbf{FT}), and prompt-based fine-tuning (\textbf{P+FT}),and ID fine-tuning (\textbf{FT-ID}). We report average \textit{Applicability} ($\mu_{F_1}$), \textit{Reliability} ($\mu_{\tau}$), \textit{Stability} ($\sigma_{F_1}, \sigma_{\tau}$). The best performance within one LM is \underline{underlined}, overall is marked in \textbf{bold}, and $\dagger$ indicates that OOD surpasses ID.}
  \label{tab:mainresults}
\end{table*}

\section{Results}\label{sec:results}
\begin{figure}[t]
  \centering
  \includegraphics[width=0.48\textwidth]{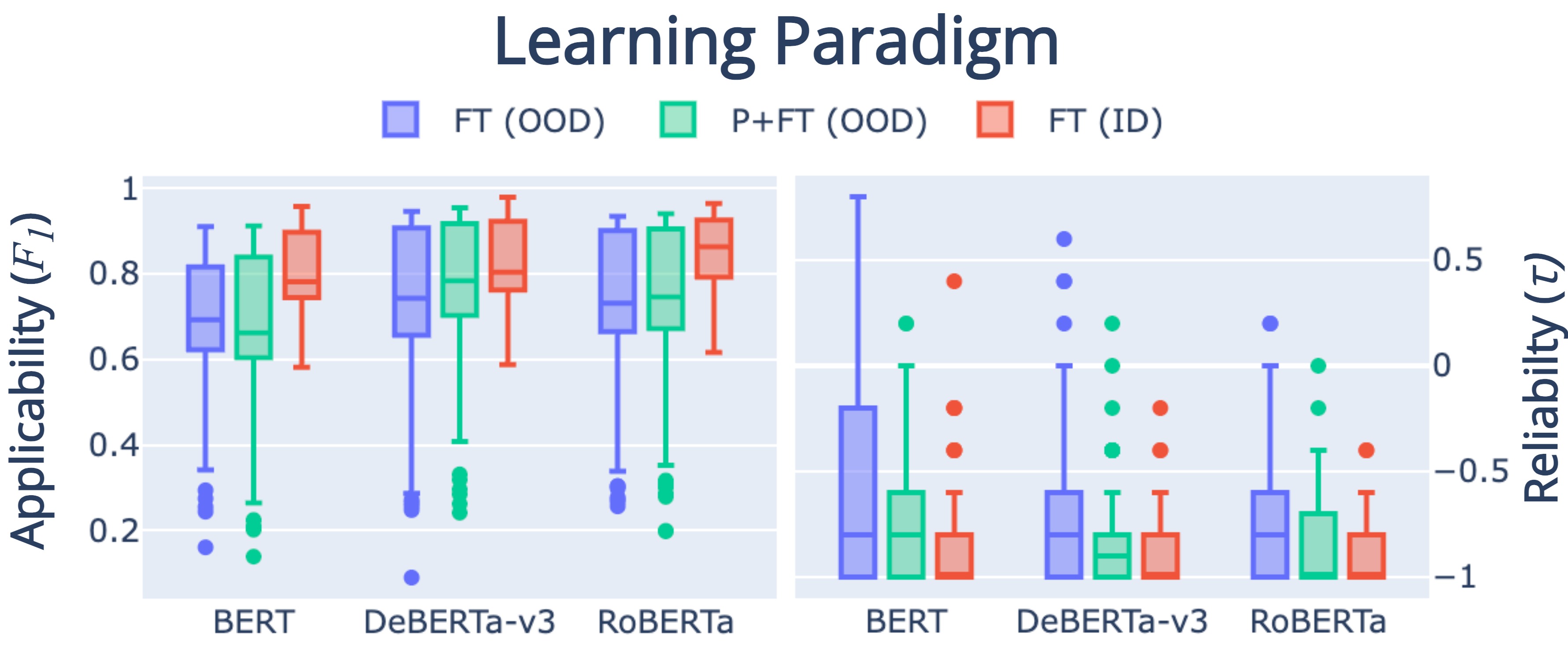}
  \caption{Aggregated results of ID and OOD vanilla fine-tuning (FT-ID and FT) and OOD prompt-based fine-tuning (P+FT) across eleven tasks (\autoref{sec:ood-tasks}) for \textit{Applicability} ($F_1$), \textit{Reliability} ($\tau$), and \textit{Stability} (deviation of $F_1$ and $\tau$).
  %We see for all models a lower \textit{Applicability},lower \textit{Reliability}, and a lower \textit{Stability}.
  }
  \label{fig:stability}
\end{figure}

This section reports results on a detailed  (\autoref{tab:mainresults}) and aggregated level (\autoref{fig:stability}) and discusses \textit{\textbf{six key findings}}.

\paragraph{i) Generalization flaws and the efficacy of prompt-based fine-tuning.}
The aggregation of the comprehensive evaluation (\autoref{fig:stability}) reveals crucial generalization flaws of OOD fine-tuning (blue). 
Compared to ID fine-tuning (red), it provides a lower \textit{Applicability} ($F_1$ score), optimization (loss) and performance ($F_1$ score) are less aligned (lower \textit{Reliability}), and measurements are less stable across different seeds and folds (\textit{Stability}).
In particular, we see this misalignment of loss and performance - a violation of a fundamental generalization assumption - crucially affects vanilla fine-tuning's (FT) degraded OOD generalization capabilities.
Turning to prompt-based fine-tuning (P+FT, green), it partially overcomes these flaws. 
Paired with DeBERTa-v3 and RoBERTa, it achieves higher absolute performance (\textit{Applicability}), a better \textit{Reliability}, and fewer deviations regarding data and randomness (\textit{Stability}).

\paragraph{ii) Superiority of DeBERTa-v3.}
Next, we focus on the detailed results (\autoref{tab:mainresults}) to compare the different LMs.
Overall, we note the superior performance of DeBERTa-v3 compared to RoBERTa and BERT for all learning paradigms across all tasks.\footnote{
These findings extend to ID scenarios — see Appendix \autoref{subsec:id-results}.} 
In particular, when paired with prompt-based fine-tuning (P+FT), DeBERTa-v3 provides $3.3$ better \textit{Applicability} ($\mu_{F_1}$), $2.9$ better \textit{Reliability} ($\mu_{\tau}$), and similar \textit{Stability} with $-0.4$ ($\sigma_{F_1}$) and $+0.3$ ($\sigma_{\tau}$) than RoBERTa with P+FT. 
Moreover, we see DeBERTa-v3 with P+FT outperforms FT in ten out of eleven tasks and reaches ID performance for two tasks (\textit{arg-sim} and \textit{review}).

\paragraph{iii) Label differences cause significant generalization gaps.}
\autoref{tab:mainresults} reveals significant generalization gaps between OOD (FT and P+FT) and ID results (FT-ID) for \textit{stance} and \textit{entail}.
These difficulties correlate with their previously identified label differences between train and test instances based on their high KL divergences (\autoref{tab:task-details}). 
These generalization issues are also visible when we compare linear probing (LP) and cloze prompting (P) for \textit{stance} and \textit{entail} with other tasks.
Since these two paradigms largely evaluate the pre-trained capabilities of LMs, we expect a big gap between them and full LM tuning paradigms (FT and P+FT) when they are done without significant generalization flaws. 
However, this gap is smaller for \textit{stance} and \textit{entail} than for other tasks, indicating that PF and P+FT exhibit higher generalization problems for these two tasks. 
%Further, a small gap between LP and P hints at potential generalization flaws since LP involves some learning. 
%Again, the average gap of around $3.5pp$ between LP and P for \textit{stance} and \textit{entail} compared to other tasks (up to $16.4pp$) further confirms these flaws.
Still, we see again that P+FT partially overcomes such generalization flaws and provides, paired with DeBERTa-v3, improvements of $3.4$ (\textit{stance}) and $4.6$ (\textit{entail}) compared to FT.

\begin{figure}[t]
 \centering
 \includegraphics[width=0.48\textwidth]{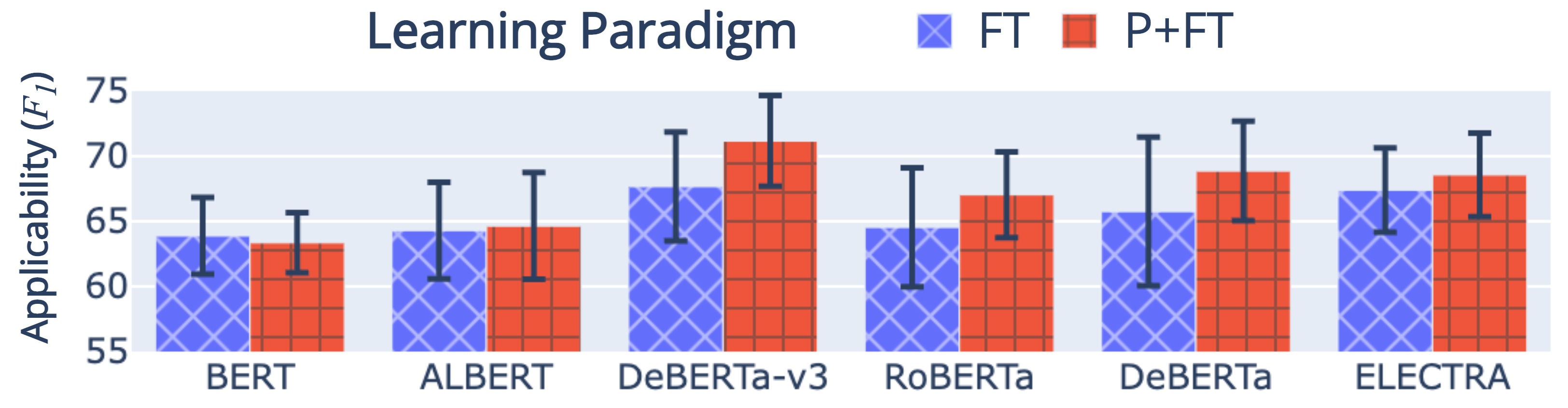}
 \caption{Average \textit{Applicability} of comparing various LMs tuned on the English-only tasks using vanilla fine-tuning (FT) or prompt-based fine-tuning (P+FT).}
 \label{fig:model_comparision}
\end{figure}

\paragraph{iv) Pre-training influences the success of prompt-based fine-tuning.}
Next, we compare the gap between vanilla fine-tuning (FT) and prompt-based fine-tuning (P+FT) for three additional base-sized LMs to better understand the efficacy of DeBERTa-v3 paired with P+FT.
In particular, we focus on its design properties like token-only pre-training objective, disentangled attention (DA), ELECTRA-style training, and extensive vocabulary. 
To determine which design choice of DeBERTa-v3 has the greatest impact on its superior OOD performance, we evaluate additional LMs on the English-only tasks involving topic and domain shifts to test these properties (\autoref{fig:model_comparision}).
First, we found that DeBERTa(-v3), RoBERTa, and ELECTRA benefit more from prompt-based fine-tuning when pre-trained with token-only objectives (like masked language modeling or replaced token detection).
In contrast, LMs such as BERT or ALBERT, trained with additional sentence objectives like next sentence prediction or sentence order prediction, exhibit minor gains or perform worse with P+FT than FT.  
Second, we do not find DeBERTa-v3 gains from ELECTRA-style pre-training, as the FT vs. P+FT gap is more pronounced for DeBERTa than ELECTRA itself. 
Third, DeBERTa (with DA) performs better than RoBERTa (without DA) on both FT and P+FT. 
DeBERTa's disentangled attention (DA) mechanism impacts its superior OOD performance since both models are pre-trained on the same datasets with masked language modeling.  
Finally, we see DeBERTa-v3's extensive vocabulary (120k tokens) as another factor in its success, as it outperforms its ancestor (DeBERTa) with 50k tokens. 
These results show how pre-training crucially shapes LMs differently beyond their performance on downstream applications \citep{wang-etal-2018-glue}.
We see these insights to be well aligned with other work, particularly %focusing on
in the examination of 
how the internal representations of LMs vary among different pre-training setups \citep{waldis-etal-2024-dive}.

%Overall, we see P+FT making use of other aspects of LMs as the inter-model ranking differs between FT and P+FT. 
%For example, ELECTRA performs 2nd best for FT but 3rd best for P+FT.

\begin{figure*}[t]
 \centering
 \includegraphics[width=0.94\textwidth]{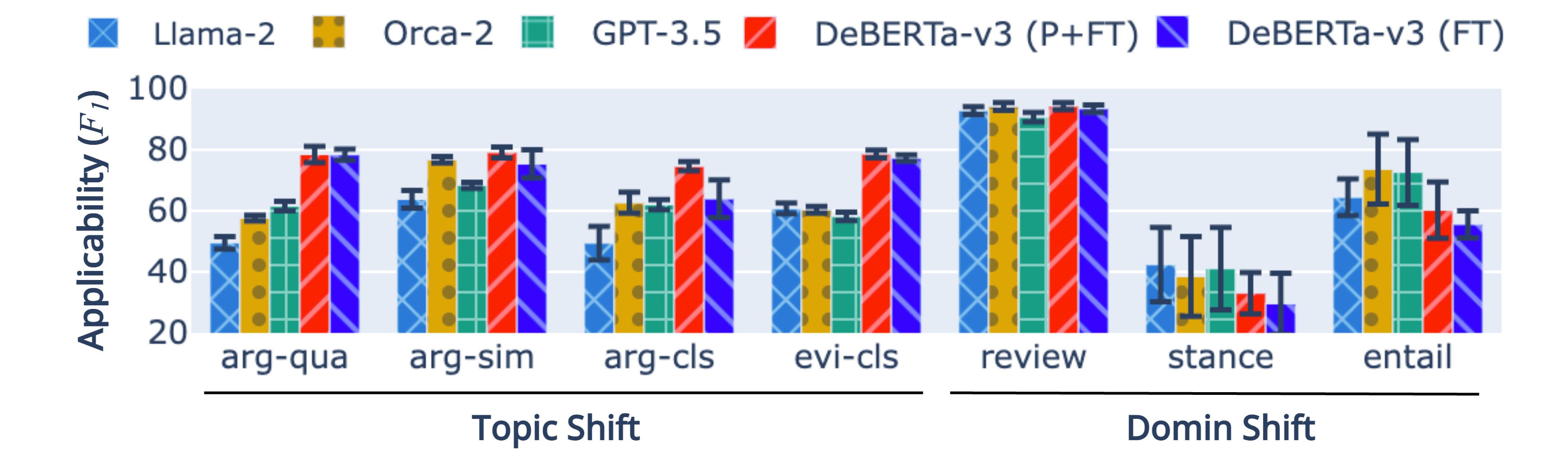}
 \caption{Comparison of ICL using ChatGPT, and DeBERTa-v3 using vanilla fine-tuning (FT) and prompt-based fine-tuning (P+FT).}
 \label{fig:icl}
\end{figure*}

\begin{table*}[t]
\centering
  \setlength{\tabcolsep}{5pt}
  \resizebox{0.90\textwidth}{!}{%
    \begin{tabular}{l|cccc|ccc|cc}
    \toprule
       & \bf arg-qua & \bf arg-sim & \bf arg-cls & \bf evi-cls & \bf review & \bf stance & \bf entail & \bf $\uparrow$ \textit{Applicability} & \bf $\downarrow$ \textit{Reliability} \\

    \midrule

{\textbf{P+FT}}\textsubscript{\text{DeBERTa-v3}} &  78.5 &  79.1 &    74.6 &   78.6 &  94.2 &  33.0 &    60.2 & 71.2$\pm$\small{1.3} &  \textbf{-84.4}$\pm$\small{8.3} \\
{\textbf{+P-Tuning}} & 56.3 &  54.9 &    38.1 &   54.7 &  53.7 &  \textbf{33.7} &    43.5 & 47.8$\pm$\small{1.2} & -22.0$\pm$\small{20.3}\\
{\textbf{+Prompt-Tuning}} &  54.8 &  54.0 &    38.2 &   54.6 &  53.4 &  32.3 &    43.2 & 47.2$\pm$\small{\textbf{0.7}} &  -6.0$\pm$\small{30.5} \\
{\textbf{+LoRA}} & 78.1 &  78.8 &    73.4 &   77.9 &  94.9 &  33.1 &    60.8 & 71.0$\pm$\small{1.0} &  -75.5$\pm$\small{\textbf{4.9}}\\\midrule
{*\textbf{P+FT}}\textsubscript{\text{DeBERTa-v3 (300m)}} & 81.4 & 80.0 & \textbf{78.7} &\textbf{ 79.8} & 95.3 & 31.2 & \textbf{62.6} & 72.7$\pm$\small{1.2} & -78.1$\pm$\small{8.0} \\
{*\textbf{P+FT}}\textsubscript{\text{T5 (3b)}} & 79.6 & 80.6 & 75.7 & 76.5 & \textbf{95.7} & 26.6 & 56.6 & 70.2$\pm$\small{0.8} & -73.1$\pm$\small{17.7} \\
%{\textbf{P+FT}}\textsubscript{\text{TK-Instruct (3b)}} & 81.4 & 82.0 & 78.7 & 79.0 & 95.6 & 26.0 & 55.6 & $71.2\pm0.8$ & $-79.3\pm8.4$ \\
{*\textbf{P+FT}}\textsubscript{\text{Flan-T5 (3b)}} & \textbf{81.8} & \textbf{82.3} & 78.5 & 79.3 & 96.3 & 31.0 & 62.4 & \textbf{73.1}$\pm$\small{1.0} & -75.0$\pm$\small{22.2} \\
%{\textbf{P+FT+LoRA}}\textsubscriptit{PYTHIA}&  67.4 &  52.5 &    52.6 &   75.2 &  94.3 &  25.9 &    44.2 & $58.9\pm1.2$ \\
%{\textbf{P+FT+LoRA}}\textsubscriptit{T5} & 81.9 &  81.0 &    78.3 &   78.8 &  95.7 &  23.6 &    56.9 & $70.9\pm1.2$ \\
%{\textbf{P+FT+LoRA}}\textsubscriptit{TK-Instruct} &  81.4 &  82.0 &    78.0 &   79.0 &  95.6 &  26.0 &    55.6 & $71.1\pm1.2$ \\
%\cdashlinelr{1-10}
%\textbf{ICL}\textsubscript{ChatGPT} &  67.9 &  71.3 &    65.2 &   54.4 &  94.7 &  45.8 &    72.1 & $67.3\pm0.0$ & -\\
    \bottomrule
    \end{tabular}
  }
  \caption{Comparison of full-parameter to efficient training using DeBERTa-v3 (rows one to four) and large LMs using LoRA (*) in rows five to seven. Best performance is marked in \textbf{bold}.}
  \label{tab:peft}
\end{table*}

\paragraph{v) No free lunch for in-context learning or gradient-based methods.}

We show in \autoref{fig:icl} results of evaluating English-only tasks using in-context learning (\textbf{ICL}) with GPT-3.5 (turbo), Llama-2-chat (70B), and Orca-2 (13B).\footnote{Please find details in the Appendix (\autoref{subsec:icl-setup}).}
Comparing these LMs for average \textit{Applicability}, notably Orca-2 ($66.2$) outperforms GPT-3.5 ($64.9$) and Llama-2 ($60.4$).
We see this strongly related to the reasoning-oriented pre-training of Orca-2.
Moreover, ICL does not reach the average performance level of the best gradient-based (FT and P+FT) approach based on DeBERTa-v3.
However, we note the superiority of ICL in scenarios involving heavy domain shifts, particularly in cross-dataset tasks, such as \textit{stance} and \textit{entail}, where heavy differences between train and test label distribution (high \textbf{KL} divergence) cause substantial generalization flaws for gradient-based learning methods.
These flaws are visible when comparing the gap between P and P+FT. 
Since we tune the LM for P+FT, we expect a significant gap for successful generalization. 
However, these gaps are relatively small for \textit{stance} and \textit{entail} - from \autoref{tab:mainresults} $+6.5$ for \textit{stance} and $+5.6$ for \textit{entail} with DeBERTa-v3 compared to $+34.3$ for \textit{arg-cls}.
In addition, there is no clear gain of using P+FT %over
for the relatively easy and popular sentiment analysis task (\textit{review}).
Due to its popularity, we assume this task is well covered in the enormous pre-training corpus of large LMs - such as GPT-3.5. 
In contrast, we note the superiority of P+FT for topic shift scenarios, which predominantly involve challenges of a semantic nature and moderate label distribution differences. 

\paragraph{vi) Few parameters are enough for competitive performance and allow to scale to larger LMs.}
Next, we compare the performance of different parameter-efficient tuning strategies with full model tuning.
As shown in rows two to four in \autoref{tab:peft}, we see LoRA with $r=4$ outperforms P-Tuning and Prompt-Tuning on most tasks.
Further, it performs on par with full-parameter tuning (first row) regarding \textit{Applicability}, provides better \textit{Reliability}, but degraded \textit{Stability}.
Further experiments considering bigger LMs show that LoRA allows their efficient use for OOD scenarios. 
Precisely, the large version of DeBERTa-v3 with 300 million provides $1.7$ higher \textit{Applicability} and $2.6$ better \textit{Reliability} than the base version (86m). 
Simultaneously, this scaling effect does not continue.
T5 or Flan-T5, with three billion parameters, seem to be generally more affected by random seeds and different folds (\textit{Stability}) without apparent \textit{Applicability}.
From these observations, OOD fine-tuning still leaves a large potential of LMs unexploited, while larger LMs improve the performance but introduce new \textit{Stability} flaws.

%% file: parts/5-Analysis.tex
\section{Analysis}\label{sec:analysis}
Next, we focus on \textit{arg-cls}, where we observe prominent differences, and discuss \textit{\textbf{four aspects}} differentiating learning paradigms.

\begin{figure*}[t]
 \centering
 \includegraphics[width=0.95\textwidth]{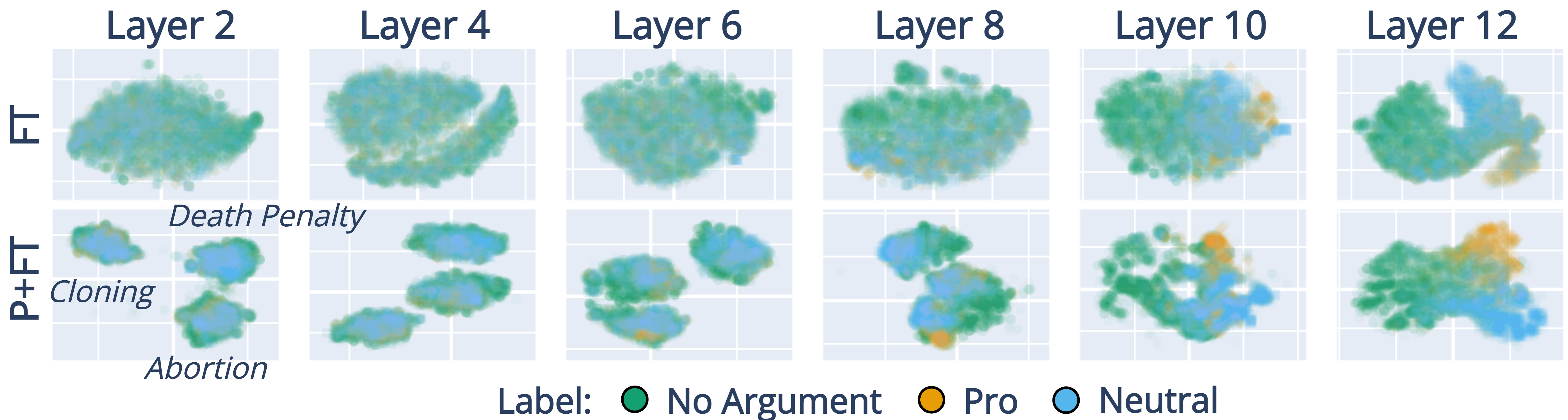}
 \caption{Overview of the T-SNE reduced embeddings of the \textit{CLS} token for FT and \textit{MASK} P+FT for every second layer where instance labels are colorized.}
 \label{fig:embeddings}
\end{figure*}

\begin{figure}[t]
 \centering
 \includegraphics[width=0.48\textwidth]{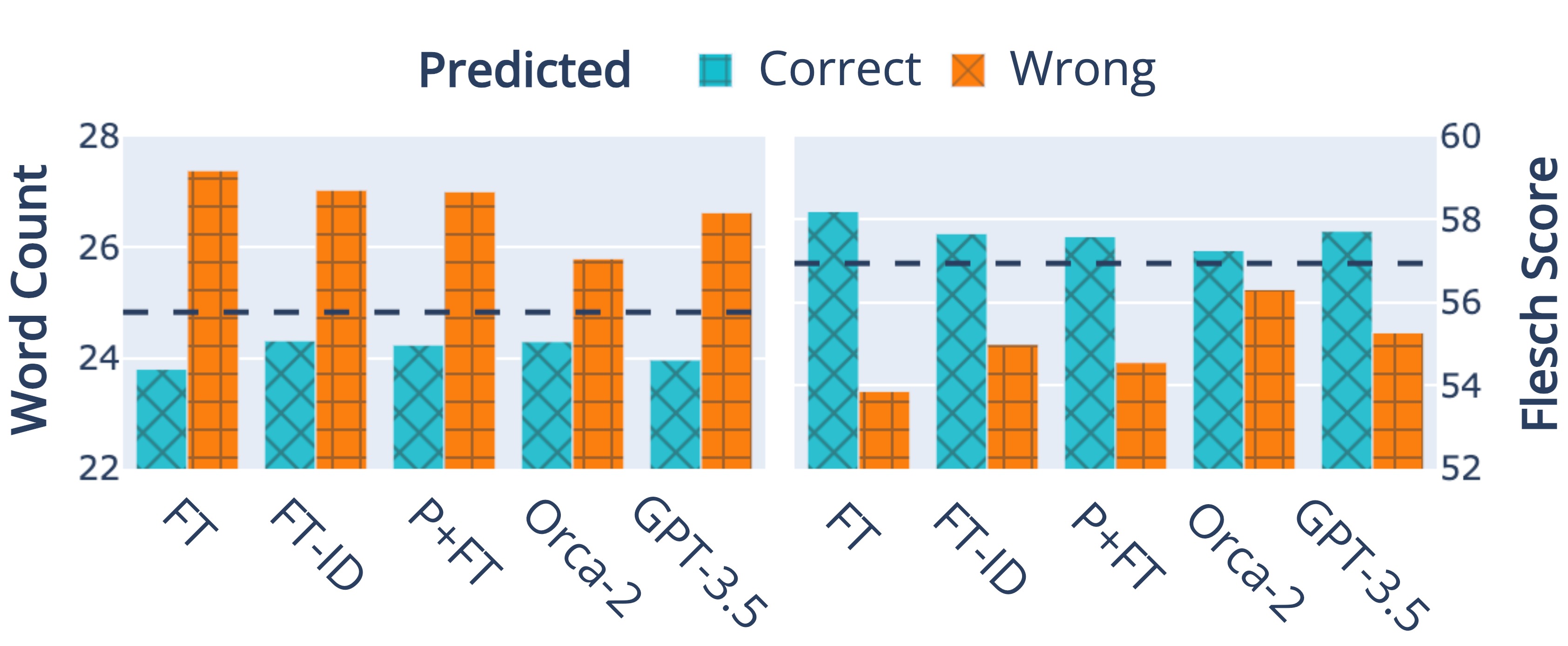}
 \caption{Average word count and input complexity (Flesch score) for correct and wrong predictions for DeBERTa-v3 with ID and OOD vanilla fine-tuning (FT-ID and FT), prompt-based fine-tuning (P+FT) and in-context learning (ICL) using Orca-2, and GPT-3.5.}
 \label{fig:word_count-Flesch}
\end{figure}

\paragraph{i) The bias regarding surface features.}
We show in \autoref{fig:word_count-Flesch} average word counts and input complexities of test instances for ID and OOD vanilla fine-tuning (FT-ID and FT) and prompt-based fine-tuning with (P+FT) using DeBERTa-v3 and in-context learning (ICL) with Orca-2 and GPT-3.5.
LMs predict shorter and more complex instances (higher Flesch score) more likely correct, and vice versa for wrong ones - compared to the dataset average (dashed line). 
However, P+FT exhibits less bias on surface correlations than FT and shows similar patterns as FT-ID, hinting at the superior abilities of P+FT.
In contrast, ICL predictions, in particular of Orca-2, are less biased for both surface features. 
Still, deviations from the dataset average suggest fundamental bias in such features.

\paragraph{ii) P+FT provides more prediction confidence and relies less on surface features. }
%While we previously considered LMs as black boxes, we next focus on the inners of LMs tuned using FT and P+FT.
From \autoref{tab:lm-inners}, P+FT provides higher average confidence (defined as the logit of the predicted label) than FT and a similar one as ID fine-tuning (FT-ID).
Thus, we assume P+FT is less confused by the distribution shift, which is also visible in the lower correlation between confidence and surface features (Flesch score or word counts) than FT. 
For example, FT seems less confident when the input is longer and more complex. 

\paragraph{iii) Prompt-based fine-tuning considers input differently.}
Next, we analyze how LMs attribute to the input tokens. 
We follow \citet{kobayashi-etal-2020-attention} and calculate the attribution of a token using the attention and the norm of the token embeddings. 
FT-ID and FT have higher average attributions than P+FT, indicating that token attributions are more evenly distributed since they are normalized using the Euclidean norm within a given input sentence. 
This is already visible when comparing the token attribution before fine-tuning (\textit{raw} vs. \textit{P+raw}).
Apparent differences are also visible when we compare how attributions of correct or wrong predicted instances differ. 
While P+FT shows maximum 0.4 differences (P+FT), this rises to 1.0 for FT-ID.
With these results, we assume LMs applied in prompt-based or vanilla fine-tuning fundamentally differ in how inputs are processed. 

\paragraph{iv) P+FT retains more semantic information.}
\autoref{fig:embeddings} visualizes the layer-wise embeddings of the classification proxy tokens - \textit{CLS} for FT and \textit{MASK} for P+FT.
It shows that P+FT retains more semantic information (about topics) until the last layers, while FT eliminates them across all layers during training.
Hinting, again, at substantial differences between FT and P+FT.
\begin{table}[t]
\centering
  \setlength{\tabcolsep}{2pt}
  \resizebox{0.48\textwidth}{!}{%
    \begin{tabular}{l|ccc|cc}
    \toprule
     & \textbf{FT-ID} & \textbf{FT} & \textbf{P+FT} & \textbf{raw} & \textbf{P+raw} \\ \midrule
    Average Confidence & 97.6 & 95.9 & 97.8 & - & - \\  
    Confidence$\times$Flesch & 5.1 & 8.6 & 4.1 & - & - \\    
    Confidence$\times$Word Count& -10.3 & -13.2 & -6.3 & - & - \\\midrule    
    Average Attribution & 16.2 & 15.5 & 13.0 & 16.3 & 13.2 \\  
    
    Correct Attribution & 16.4 & 15.8 & 13.1 & - & -\\  
    Wrong Attribution & 15.2 & 14.9 & 12.7 & - & - \\  
    %Attribution$\times$Certainty& 8.8 & 11.0 & 11.4 & 4.9 & - & - \\   
    %Split Tokens Attribution & - & 14.5 & -1.4 & 17.3 & 0.2 & 5.0 \\   
    % \> correct & 16.4 & 15.8 & 14.2 & 13.1 & - & -\\  
    % \> wrong & 15.2 & 14.9 & 14.4 & 13.4 & - & - \\ 
    %Label Tokens Attribution & - & 4.1 & 1.0 & 1.8 & 0.4 & 1.8 \\    
    % \> correct & 16.9 & 16.9 & 16.3 & 15.8 & - & -\\  
    % \> wrong & 16.8 & 16.8 & 16.0 & 16.4 & - & - \\ 
    %Attribution$\times$Label& 2.3 & -1.7 & -1.9 & -4.2 & -2.0 & -3.3 \\   
    %Attribution$\times$Split+Label& 3.1 & 6.7 & 5.2 & 10.2 & 5.3 & 1.3 \\     
    \bottomrule
    \end{tabular}
  }
  \caption{Analysis and correlation ($\times$) of the prediction confidence and token attribution for DeBERTa-v3. \textit{raw} and \textit{P+raw} provide results of the solely pre-trained LM.}
  \label{tab:lm-inners}
\end{table}

%% file: parts/7-Discussion.tex
\begin{comment}
\section{Discussion}
We exhaustively evaluated a variety of OOD tasks with different LMs and learning paradigms.
These results and the in-depth analyses shed light on crucial flaws of OOD generalization and how prompt-based fine-tuning can partially overcome them.
Different from previous work \cited{yuan2023revisiting}, we do not find substantial grounds to prioritize in-context learning over gradient-based learning for OOD scenarios generally. 
Instead, our work emphasizes the heterogeneity of OOD generalization.
This diversity requires fine-grained examination, like a detailed analysis of specific distribution shifts, to decide which learning paradigm to prefer.
Specifically, in-context outperforms gradient-based learning for domain shifts encompassing more structural challenges such as label divergence. 
Opposedly, we found the superior performance of gradient-based learning when expecting more semantic shifts, as in topic shifts.

Furthermore, additional factors are relevant to have in mind when analyzing these results. 
Involving the out-of-the-box application and flexibility of ICL, the computational efficiency of gradient-based approaches \citep{mosbach-etal-2023-shot, sheng2023s, yu2023open}, and the issue of task contamination in large LMs \citep{li2023task}.

\end{comment}

\section{Conclusion}
This work marks the most extensive study to date addressing the heterogeneous types of OOD scenarios in CA by systematically evaluating different OOD types. 
We evaluate a multiplicity of LMs and learning paradigms on eleven CA tasks.
With this extensive evaluation, we shed light on the challenges of having diverse covariant distribution shifts in CA.
In addition, we provide takeaways of general relevance, such as the superiority of ICL for domain shifts, where gradient-based learning fails to %generalize due to heavy label divergences between train and test data.
generalize effectively due to significant label discrepancies between the training and testing data. 
In contrast, gradient-based learning surpasses ICL  when generalization across significant semantic differences is required, like in cases of topic shifts. 
%While being particularly relevant for CA field, it also provides numerous insights of general relevance when facing. 
%Particully
%We show, amongst other findings, that in-context and gradient-based methods excel for different types of OOD scenarios. 
%This analysis highlights the importance of considering the heterogeneity of OOD scenarios to draw profound conclusions. 
With the rise of larger LMs, systematic evaluation of distribution shifts becomes even more important,  necessitating the consideration of additional factors such as computational efficiency, task complexity, and data contamination.
Finally, our findings highlight the untapped potential of base-sized models, which points towards a need for further advancements in gradient-based learning paradigms.

\section*{Ethical Considerations and Limitations}

\subsection{Higher Input Length}
By embedding the input into a prompt, we sacrifice potential input tokens. 
Since the used tasks have relatively short inputs, this is not crucial for this work.
However, this can be an essential limitation for other tasks when inputs get longer.

\subsection{Efficiency}
We always refer to efficient fine-tuning when discussing efficient methods in this work. 
Therefore, we did not consider efficient methods to make inferences on larger LMs more feasible. 
We see this as another crucial and essential aspect of real-world applications.
Simultaneously, we think performance and efficiency will alternate in the future. 
Therefore, we keep that for future work.

\subsection{Large Language Models}
We show the competitive performance of ChatGPT compared to gradient-based approaches by only relying on four demonstrations and without any tuning. 
Simultaneously, we need to assume that the pre-training corpus of ChatGPT leaks crucial aspects - like broadly covers controversially discussed topics like \textit{Nuclear Energy} or includes instances of popular datasets (like \textit{RTE}  \citep{wang-etal-2018-glue} or \textit{SemEval2016} \citep{mohammad-etal-2016-semeval}) word-by-word.
When we have in mind that we use OOD to verify generalization capabilities required for upcoming scenarios, we need to examine the performance of ChatGPT carefully and whether it was able to learn the task or just remembered some semantic aspects of the pre-training.

\section*{Acknowledgements}
We thank  Cecilia Liu, Thy Thy Tran, and Kexin Wang for their valuable feedback.
Andreas Waldis has been funded by the Hasler Foundation Grant No. 21024.
Yufang Hou is supported by the Visiting Female Professor Programme from TU Darmstadt.

%% file: parts/9-Appendix.tex
\clearpage
\appendix

\section{Additional Details of the Experiments}\label{sec:experiment-detail}

\subsection{Training Setup}
For all our experiments, we use NVIDIA RTX A6000 GPUs with CUDA (11.7), python (3.8.10), transformers (4.28.0), PyTorch (1.13.1), and openprompt (1.0.1).

\subsection{Hyperparameters}\label{subsec:hyperparameters}
We use for the experiments fixed hyperparameters; AdamW \citep{adamW2019} as optimizer; a batch size of 16; a learning rate of 0.00002; a dropout rate of 0.1; a warmup rate of 10\% of the steps; random seeds: $[0,1,2]$.
In the case of parameter-efficient tuning, we use a learning rate of a learning rate of 0.0002.
Moreover, we use the following tags from the huggingface model hub:

\begin{itemize}
    \item \href{https://huggingface.co/albert-base-v2}{\texttt{albert-base-v2}}
    \item \href{https://huggingface.co/bert-base-uncased}{\texttt{bert-base-uncased}}
    \item \href{https://huggingface.co/aajrami/bert-mlm-base}{\texttt{aajrami/bert-mlm-base}}
    \item \href{https://huggingface.co/microsoft/deberta-base}{\texttt{microsoft/deberta-base}}
    \item \href{https://huggingface.co/microsoft/deberta-v3-base}{\texttt{microsoft/deberta-v3-base}}
    \item \href{https://huggingface.co/roberta-base}{\texttt{roberta-base}}
    \item \href{https://huggingface.co/google/electra-base-discriminator}{\texttt{google/electra-base\-discriminator}}
    \item \href{https://huggingface.co/t5-3b}{\texttt{t5-3b}}
    \item \href{https://huggingface.co/google/flan-t5-xl}{\texttt{google/flan-t5-xl}}
    \item \href{https://huggingface.co/TheBloke/Llama-2-70B-Chat-AWQ}{\texttt{TheBloke/Llama-2-70B-Chat-AWQ}}
    \item \href{https://huggingface.co/TheBloke/Orca-2-13B-AWQ}{\texttt{TheBloke/Orca-2-13B-AWQ}}
    %\item \href{https://huggingface.co/EleutherAI/pythia-2.8b}{\texttt{EleutherAI/pythia-2.8b}}
\end{itemize}

\subsection{Fold Composition}\label{subsec:appendix:folds}
With our evaluation, we want to cover a given dataset fully. 
Therefore, we conduct a multi-folded evaluation that covers every instance of the dataset once in one of the tests splits $X_{\text{test}}$.
For a fair comparison, we use the same number of folds for OOD and ID and synchronize their dimension, i.e., train, dev, and test split of the first fold have the same number of instance for OOD and ID. 

\begin{figure}[]
  \centering
  \includegraphics[width=0.48\textwidth]{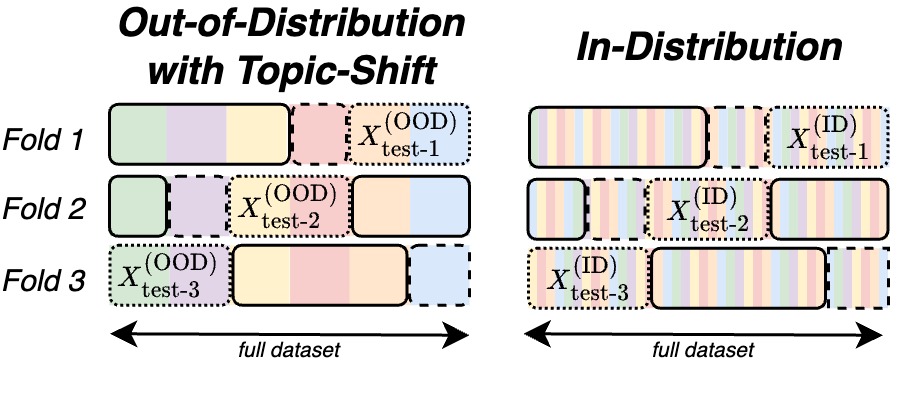}
  \caption{Example of the composition of the different folds when we target the topic shift of a dataset with three folds. Distinct topics are colorized, while solid, dashed, or dotted lines indicate train, dev, and test splits.}
  \label{fig:folds}
\end{figure}

We show with \autoref{fig:folds} an example of a dataset with a topic shift.
We colorize topics and indicate train, dev, and test splits with solid, dashed, and dotted lines.
First, we sort all dataset instances according to their assigned topic for OOD while we randomly shuffle them for ID. 
Then we compose the test splits $X_{\text{test}}^{\text{(OOD)}}$ and $X_{\text{test}}^{\text{(ID)}}$in a way to cover every instance exactly once.
Next, we form the train and dev splits by randomly distribution the left-over topics (OOD) or instances (ID) for all folds.
When composing these splits, we compose the ID splits to match the respective OOD splits of the same fold.
For example considering the first fold, the splits $X_{\text{train-1}}^{\text{(OOD)}}$, $X_{\text{dev-1}}^{\text{(OOD)}}$, and $X_{\text{test-1}}^{\text{(OOD)}}$ have the same number of instances as the splits  $X_{\text{train-1}}^{\text{(ID)}}$, $X_{\text{dev-1}}^{\text{(ID)}}$, and $X_{\text{test-1}}^{\text{(ID)}}$.

Based on the number of unique distribution shift properties (topics, domains, or languages), we use the different number of folds to distribute these properties as even as possible across the different test splits $X_{\text{test}}^{\text{(OOD)}}$.
Therefore, we use whenever possible a three-folded setup.
However, when the number of distribution properties is equal to four (i.e., four domains), we conduct a four-folded evaluation.
Please find this concrete number of folds per task in the source code.

\subsection{Dataset Details}\label{subsec:dataset-examples}

\begin{figure}[t]
 \centering
 \includegraphics[width=0.48\textwidth]{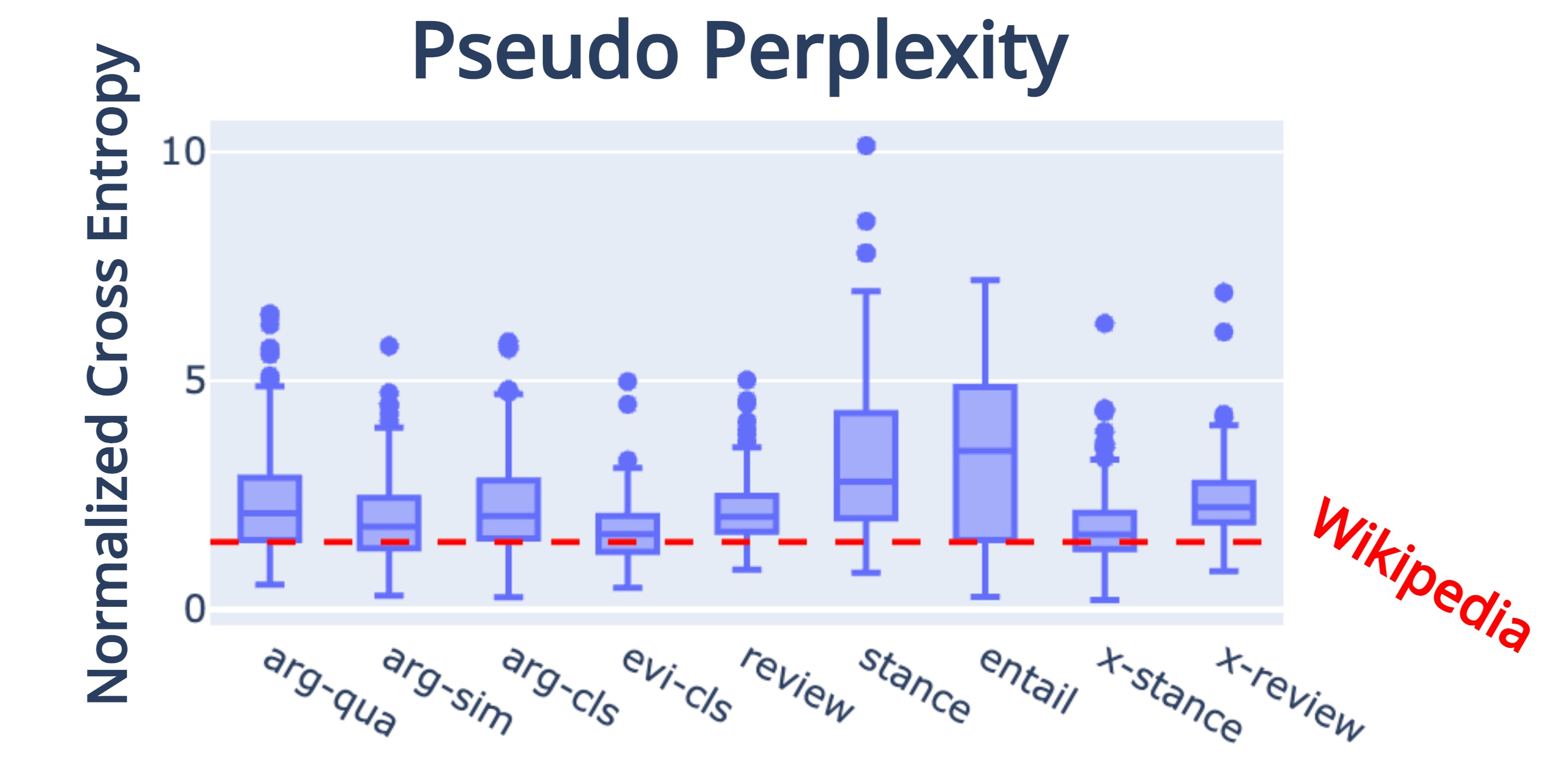}
 \caption{Pseudo perplexity of the selected tasks compared to pre-training data from Wikipedia (red line).}
 \label{fig:pseudo-perplexity}
\end{figure}

As a part of this work, we propose eleven different OOD classification tasks based on 13 different datasets.
In the following, we provide additional details.
\autoref{tab:datasets} shows an overview of these tasks and examples for every task. 
Furthermore, we show in \autoref{fig:pseudo-perplexity} how these task examples diverge from the LMs' pre-trained textual understanding based on Wikipedia, which is a major pre-training dataset for BERT, RoBERTa, and DeBERTa.  
Specifically, we compute the pseudo perplexity \citep{salazar-etal-2020-masked}, determined as the cross-entropy of each token, for 500 randomly chosen instances per task. 
For English tasks, \texttt{bert-base-uncased} is used, while \texttt{bert-base-multilingual-uncased} is deployed for multilingual tasks.
We then compare the averaged cross-entropy of these tasks (refer to \autoref{fig:pseudo-perplexity}) against 500 randomly selected samples from the Wikipedia pre-training corpus \citep{devlin-etal-2019-bert}. 
The comparison indicates a notable divergence in the chosen tasks from a common pre-training dataset Wikipedia, except for \textit{evi-cls}, which leverages Wikipedia data, and \textit{x-stance}, which aligns closely with Wikipedia's text genre.

Finally, \autoref{tab:prompts} lists the prompt templates used for all tasks and languages.

\begin{table*}[t]
\centering
    \setlength{\tabcolsep}{3pt}
    \resizebox{1.00\textwidth}{!}{%
    \begin{tabular}{llll}
    \toprule
    & \bf Dataset & \bf Description & \bf Distribution Shift \\
    \midrule
    \textbf{arg-qua} & \makecell[l]{Argument Quality \\ \citep{toledo-etal-2019-automatic}} & \makecell[l]{\\Choose which argument out of two has the higher quality: \\ \pattern{TOPIC}: \textit{we should ban fossil fuels} \\ \pattern{ARG-1}: \textit{fossil fuels pollute and cause a lot of diseases} \\ \pattern{ARG-2}: \textit{fossil fuel companies often have incredibly bad and dangerous working conditions} \\ \pattern{LABEL}: \pattern{ARG-1}\\ \textit{}} & \textbf{Topical} (22 topics)\\

    \textbf{arg-sim} & \makecell[l]{Argument Similarity \\ \citep{reimers-etal-2019-classification}} &  \makecell[l]{Decide whether two arguments are \pattern{similar} or \pattern{not-similar}: \\ \pattern{TOPIC}: \textit{organ donating} \\ \pattern{ARG-1}: \textit{One organ and tissue donation can save or enhance the lives of nearly 100 people} \\ \pattern{ARG-2}: \textit{By donating your organs after you die, you can save or improve as many as 50 lives} \\ \pattern{LABEL}: \pattern{similar}} & \textbf{Topical} (28 topics)\\
    
    \textbf{arg-cls} & \makecell[l]{Argument Classification \\ \citep{stab-etal-2018-cross}} &  \makecell[l]{\\Classify an argument as \pattern{pro}, \pattern{con}, or \pattern{no-argument} given a topic: \\ \pattern{TOPIC}: \textit{abortion} \\ \pattern{ARG}: \textit{Now our nonprofit really needs your help} \\ \pattern{LABEL}: \pattern{similar}\\\textit{}}& \textbf{Topical} (8 topics)\\

    \textbf{evi-cls} & \makecell[l]{Evidence Classification \\ \citep{shnarch-etal-2018-will}} & \makecell[l]{Decide whether a text is \pattern{relevant} evidence for a topic or \pattern{not-relevant} \\ \pattern{TOPIC}: \textit{we should limit executive compensation} \\ \pattern{TEXT}: \textit{On April 7, 2009, Blankfein recommended guidelines to overhaul executive compensation} \\ \pattern{LABEL}: \pattern{not-relevant} \\ \textit{}} & \textbf{Topical} (118 topics)\\ \midrule
    
    \textbf{review} & \makecell[l]{Sentiment Classification \\ \citep{DBLP:conf/acl/BlitzerDP07}} & \makecell[l]{\\Classify product review as \pattern{positive} or \pattern{negative}: \\ \pattern{DOMAIN}: \textit{dvd} \\ \pattern{REVIEW}: \textit{If you don't own this dvd ... my opinion it is the best american animated film ever released}\\\pattern{LABEL}: \pattern{positive}\\\textit{}} & \makecell[l]{\textbf{Domain}\\books, dvd, electronics, \\ kitchen \& housewares} \\
    
    \textbf{stance} & Stance Detection & \makecell[l]{Classify a text as either \pattern{pro}, \pattern{con}, or \pattern{neutral} regarding a topic: \\ \pattern{TOPIC}: \textit{climate change is a real concern} \\ \pattern{TEXT}: \textit{Be kind to the earth beneath your feet.  \#environment}\\\pattern{LABEL}: \pattern{pro}} & \makecell[l]{\textbf{Domain} \\News \citep{ferreira-vlachos-2016-emergent}\\ Debating \citep{DBLP:conf/lrec/WalkerTAAK12}\\ Social Media \citep{mohammad-etal-2016-semeval}}\\
    
    \textbf{entail} & Entailment & \makecell[l]{\\Predict whether two sentences do \pattern{entail} or \pattern{not-ential} each other: \\ \pattern{DOMAIN}: RTE \\ \pattern{SENTENCE-1}: \textit{No Weapons of Mass Destruction Found in Iraq Yet}\\\pattern{SENTENCE-2}: \textit{Weapons of Mass Destruction Found in Iraq} \\\pattern{LABEL}: \pattern{not entail}\\\textit{}} & \makecell[l]{\textbf{Domain} \\RTE \citep{wang-etal-2018-glue}\\ SciTail \citep{DBLP:conf/aaai/KhotSC18}\\ HANS \citep{mccoy-etal-2019-right}}\\\midrule

    \textbf{x-review} & \makecell[l]{Multilingual Sentiment Classification \\ \citep{prettenhofer-stein-2010-cross}} & \makecell[l]{\\Classify product review as \pattern{positive} or \pattern{negative}: \\ \pattern{DOMAIN}: \textit{books}  \\ \pattern{LANGUAE}: \textit{de}\\ \pattern{REVIEW}: \textit{Ich war vor 5 Jahren in Indien ... Ich kann dieses Buch nur empfehlen.}\\\pattern{LABEL}: \pattern{positive}\\\textit{}} & \makecell[l]{\textbf{Domain}\\books, dvd, music\\\textbf{Lingual}\\de, en, fr, jp} \\

    \textbf{x-stance} & \makecell[l]{Multilingual Stance Detection \\ \citep{DBLP:conf/swisstext/VamvasS20}} & \makecell[l]{\\Classify a text as either \pattern{favor}, or \pattern{against} regarding a given topic: \\ \pattern{TOPIC}: \textit{encomonmy}  \\ \pattern{LANGUAE}: \textit{it}\\ \pattern{TEXT}: \textit{Non penso che tale ampliamento sia necessario, né urgente.}\\\pattern{LABEL}: \pattern{against}\\\textit{}} & \makecell[l]{\textbf{Domain}\\books, dvd, music\\\textbf{Lingual}\\de, fr, it} \\

        \bottomrule
        \end{tabular}
    }
    \caption{Overview and examples of the used datasets and information about the enforced distribution shift.}
    \label{tab:datasets}
\end{table*}

\begin{CJK}{UTF8}{min}
\begin{table}[t]
\centering
    \setlength{\tabcolsep}{3pt}
    \resizebox{0.5\textwidth}{!}{%
    \begin{tabular}{ll}
    \toprule
    \bf Task &  \bf Prompt\\
    \midrule
    arg-qua &    \pattern{ARG-1} is \pattern{MASK} than \pattern{ARG-2} regarding \pattern{TOPIC}\\

    arg-sim &    \pattern{ARG-1} is \pattern{MASK} than \pattern{ARG-2} regarding \pattern{TOPIC}\\

    arg-cls &    The attitude of \pattern{ARG} is \pattern{MASK} regarding \pattern{TOPIC}\\

    evi-cls &    \pattern{TEXT} is \pattern{MASK} evidence regarding \pattern{TOPIC}\\ \midrule

    review &    The sentiment of \pattern{REVIEW} is \pattern{MASK}\\

    stance &   The attitude of \pattern{TEXT} is \pattern{MASK} regarding \pattern{TOPIC}\\
    entail &  \pattern{SENTENCE-1}? \pattern{MASK}, \pattern{SENTENCE-2}\\ \midrule

    x-stance &   \makecell[l]{
        de: Die Haltung von \pattern{ARG} ist \pattern{MASK} zu \pattern{TOPIC} \\
        fr: L'attitude de \pattern{ARG} est \pattern{MASK} envers \pattern{TOPIC} \\
        it: L'atteggiamento di \pattern{ARG}   \pattern{MASK} verso \pattern{TOPIC}
    }\\
    
    x-review &   \makecell[l]{
        de: Die Stimmung von \pattern{REVIEW} ist \pattern{MASK}\\
        en: The sentiment of \pattern{REVIEW} is \pattern{MASK} \\
        fr: Le sentiment de \pattern{REVIEW} est \pattern{MASK} \\
        jp: \pattern{REVIEW} の感情は \pattern{MASK} です
    }
    \\
        \bottomrule
        \end{tabular}
    }
    \caption{Overview of the used prompt templates for all tasks and languages for the prompt-tuning setup.}
    \label{tab:prompts}
\end{table}

\end{CJK}

\subsection{Prompt-Based Fine-Tuning}\label{subsec:prompt-based-fine-tuning}

In this paper, we adopt prompt-based fine-tuning (\textbf{P+FT}) \citep{DBLP:journals/corr/abs-2107-13586} as an alternative approach to vanilla fine-tuning (\textbf{FT}). 
Unlike FT, P+FT relies on the pre-trained masked-language modeling (MLM) head and avoids using new classification heads.

\begin{figure}[t]
    \centering
    \includegraphics[width=0.4\textwidth]{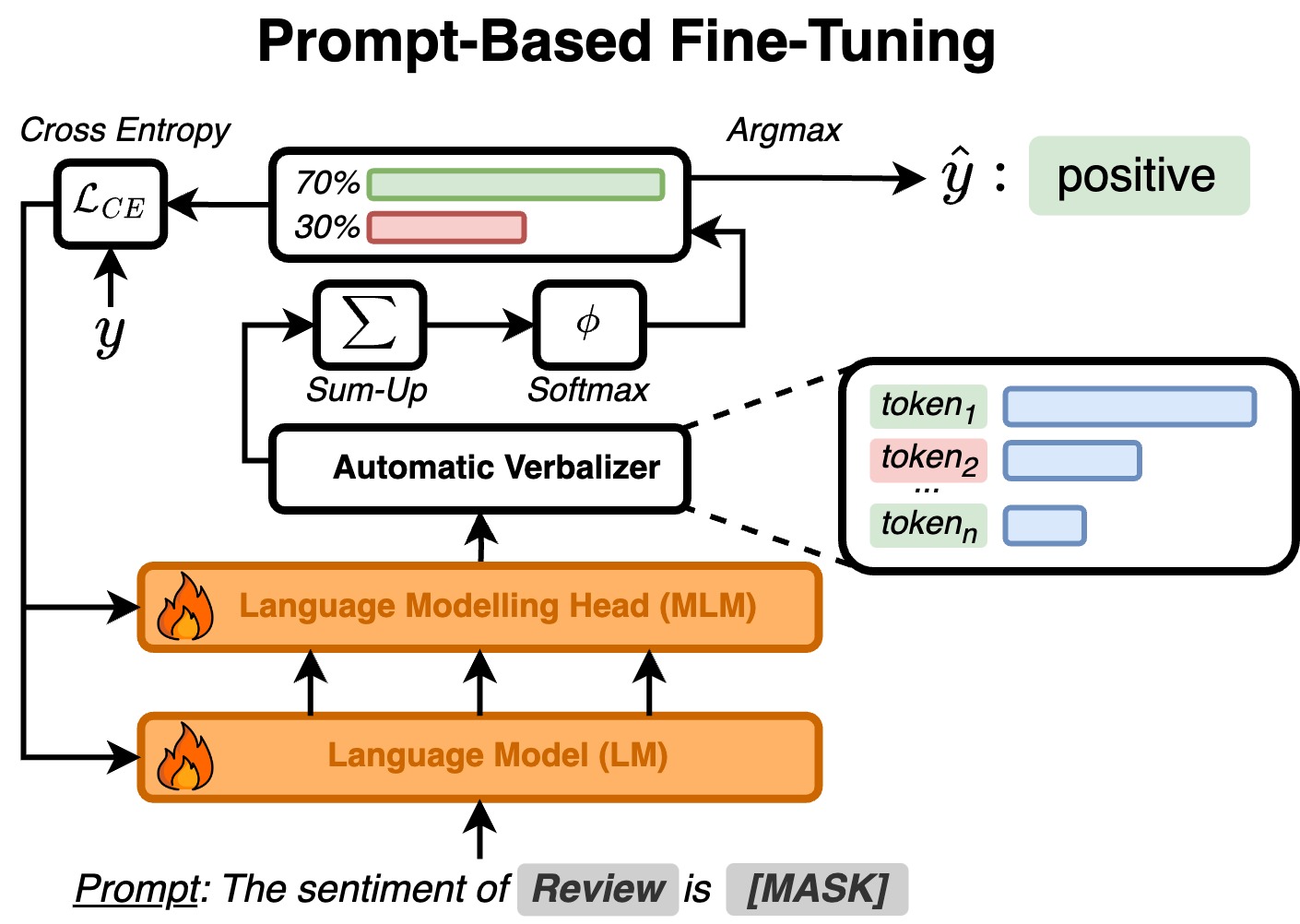}
    \caption{
    An exemplary overview of prompt-based fine-tuning (P+FT) for the sentiment classification task, following \citet{schick-schutze-2021-exploiting}:
    re-formulating the task as cloze (prompt); gathering relevant tokens (verbalizer) for the specific classes - \textit{positive} (green) or \textit{negative} (red); finding the final prediction by summing up the probability of relevant tokens per class; backpropagating the error through the MLM head and the LM.
    }
    \label{fig:overview}
\end{figure}

\paragraph{Forward Pass}
We show in \autoref{fig:overview} an exemplary overview of P+FT for sentiment analysis given two classes $K=\{\textit{positive}, \textit{negative}\}$.
In detail, we wrap the review with a cloze template and add a masking token as the prediction proxy.
Next, the LM processes this prompt and outputs the most probable tokens $T$ along with their log probabilities $L$ using the MLM head.
Then, the verbalizer selects the relevant tokens $A$ within $L$ and assigns them a class mapping - like \textit{positive} (green) or \textit{negative} (red). 
In contrast to other probing work \citep{schick-schutze-2021-exploiting, DBLP:journals/corr/abs-2303-07320}, we automatically select indicative tokens \citet{schick-etal-2020-automatically} using the likelihood ratio regarding every class $k$ in $K$ based on the train instances.
With these token-class mappings, we sum up the log-probabilities for every class $k$ in $K$ as $w_{k} = \sum_{a \in A(k)} L(a)$ and apply the softmax (\autoref{eq:softmax}) to find $\hat{y}$.

\begin{equation}\label{eq:softmax}
    \hat{y} = \argmax_{k \in K} \dfrac{exp(w_k)}{\sum_{k' \in K} exp(w_{k'})}
\end{equation}

\paragraph{Backward Pass}
While the forward pass represents the prompting paradigm (\textbf{P}), we analyze and evaluate LMs without parameter optimization. 
However, to fine-tune the LM and MLM head, we calculate the cross-entropy loss $\mathcal{L}_{CE}$ and update the weights through back-propagation.
Note that we initialized the automatic verbalizer before the training and did not update it anymore afterward.

\subsection{In-Context Learning Setup}\label{subsec:icl-setup}
As reported in \autoref{sec:results}, we evaluated ChatGPT using in-context learning (ICL).
In detail, we provide four demonstration samples from the training instances for every test instance. 
We use the templates reported in \autoref{tab:icl_prompts} for every training demonstration instance and the test instance, where we exclude the \pattern{LABEL} for the test one. 
For a fair comparison with gradient-based approaches (FT, P+FT), we allow to sample these demonstration instances from the entire training set. 
We use BM25 to calculate the similarity between the test and train instances.
Afterward, we use the top-4 most similar train instances as a demonstration for a given test instance.

\begin{table*}[t]
\centering
    \setlength{\tabcolsep}{3pt}
    \resizebox{1.0\textwidth}{!}{%
    \begin{tabular}{ll}
    \toprule
    \bf Task & \bf Prompt \\
    \midrule
    \textbf{arg-qua} & \makecell[l]{\\Given the following two arguments and the topic they cover, which one has the higher quality? Options are first or second.\\ Argument 1: \pattern{ARG-1}\\ Argument 2: \pattern{ARG-2}: \\Topic: \pattern{TOPIC} \\ Label: \pattern{LABEL}\\ \textit{}} \\

    \textbf{arg-sim} &  \makecell[l]{Are the following arguments similar regarding the given topic? Options are yes or no. \\ Argument 1: \pattern{ARG-1}\\ Argument 2: \pattern{ARG-2}: \\Topic: \pattern{TOPIC} \\ Label: \pattern{LABEL}\\} \\
    
    \textbf{arg-cls} &  \makecell[l]{\\What is the attitude of the following argument regarding the given topic?  Options are neutral, favor, or against.\\Argument: \pattern{ARG} \\Topic: \pattern{TOPIC} \\ Label: \pattern{LABEL}\\\textit{}}\\

    \textbf{evi-cls} & \makecell[l]{Corresponds the following evidence to the given topic?  Options are yes or no. \\ Evidence: \pattern{TEXT} \\ Topic: \pattern{TOPIC} \\ Label: \pattern{LABEL} \\\textit{}} \\ \midrule
    
    \textbf{review} & \makecell[l]{\\What is the sentiment of the following text?  Options are positive or negative. \\ Review: \pattern{TEXT} \\ Label: \pattern{LABEL} \\\textit{}} \\
    
    \textbf{stance} & \makecell[l]{\\What is the attitude of the following text regarding the given topic?  Options are neutral, favor, or against.\\Text : \pattern{TEXT} \\Topic: \pattern{TOPIC} \\ Label: \pattern{LABEL}} \\
    
    \textbf{entail} & \makecell[l]{\\Can we conclude an entailment from the following two texts?  Options are yes or no.\\ Text 1: \pattern{TEXT-1}\\ Text 2: \pattern{TEXT-2}: \\Topic: \pattern{TOPIC} \\ Label: \pattern{LABEL}\\\textit{}}\\\midrule

        \bottomrule
        \end{tabular}
    }
    \caption{Overview of the used prompting templates for the in-context learning setup.}
    \label{tab:icl_prompts}
\end{table*}

\section{Additional Results}\label{sec:additional-results}
In addition to results shown in the main paper (\autoref{sec:results}), we show in the following the effectiveness of P+FT for ID scenarios (\autoref{subsec:id-results}) and that other methods to prevent freshly initialized classification heads underperforms prompt-based fine-tuning (\autoref{subsec:lpft}). 

\subsection{In-Distribution Results}\label{subsec:id-results}
\autoref{tab:id-results} shows the superior performance of prompt-based fine-tuning transfers to ID scenarios.

\begin{table*}[t]
\centering
    \setlength{\tabcolsep}{3pt}
    \resizebox{1.00\textwidth}{!}{%
        \begin{tabular}{l|cccc|ccc|cc|cc}
        \toprule
       & \bf arg-qua & \bf arg-sim & \bf arg-cls & \bf evi-cls & \bf review & \bf stance & \bf entail & \bf x-stance & \bf x-review & \bf \textit{Applicability} & \bf \textit{Reliability}\\
     &  \it \small{Top.}  &  \it  \small{Top.} &  \it    \small{Top.}  &  \it  \small{Top.}   &  \it \small{Dom.}   &  \it  \small{Dom.}  &  \it   \small{Dom.}   &   \it    \small{Lang./Top.} &  \it     \small{Lang./Dom.} & $\mu_{F_1}\pm\sigma_{F1}$   &  $\mu_{\tau}\pm\sigma_{\tau}$    \\
        \midrule

{\textbf{LP}}\textsubscript{BERT}&   55.7 &   69.9 &        58.5 &     70.4 &    85.5 &    55.3 &        72.1 &      58.9 &         76.3 &  $67.0\pm0.2$ &  $-71.3\pm3.8$ \\
{\textbf{P}}\textsubscript{BERT} &   47.7 &   50.4 &        36.5 &     54.7 &     60.2 &    44.7 &        49.2 &      49.2 &         57.2 &  $48.7\pm0.0$ &    - \\
{\textbf{FT}}\textsubscript{BERT} &   87.9 &   76.4 &        67.3 &     78.9 &    90.4 &    61.1 &        93.4 &      67.6 &         87.0 &  $78.9\pm0.4$ &  $-83.7\pm6.5$ \\
{\textbf{P+FT}}\textsubscript{BERT} &   88.0 &   76.1 &        67.7 &     79.1 &    90.4 &    62.8 &        93.4 &      67.0 &         87.0 &  $79.1\pm0.3$ &  $-78.7\pm8.1$ \\ \midrule
{\textbf{LP}}\textsubscript{DeBERTa-v3} &   55.1 &   72.5 &        60.3 &     71.1 &    89.3 &    53.2 &        87.1 &      59.6 &         85.5 &  $70.4\pm0.1$ &  $-74.6\pm3.0$ \\
{\textbf{P}}\textsubscript{DeBERTa-v3}  &   55.1 &   60.5 &        41.7 &     61.5 &    63.3 &    46.1 &        57.8 &      52.2 &         53.4 &  $54.6\pm0.5$ &   - \\
{\textbf{FT}}\textsubscript{DeBERTa-v3} &   89.0 &   78.4 &        75.2 &     80.6 &    93.9 &    63.3 &        96.7 &      72.5 &         92.1 &  $82.4\pm0.4$ &  $-92.3\pm6.5$ \\
{\textbf{P+FT}}\textsubscript{DeBERTa-v3} &   90.3 &   81.5 &        78.9 &     81.5 &    94.8 &    70.1 &        96.5 &      71.4 &         92.3 &  $84.1\pm0.3$ &  $-91.0\pm7.0$ \\ \midrule
{\textbf{LP}}\textsubscript{RoBERTa} &   54.0 &   67.0 &        57.8 &     69.6 &    88.4 &    53.3 &        73.0 &      59.4 &         86.2 &  $67.6\pm0.2$ &  $-80.6\pm3.1$ \\
{\textbf{P}}\textsubscript{RoBERTa} &    54.9 &   57.1 &        45.3 &     54.7 &    79.5 &    46.3 &        55.6 &      49.2 &         76.1 &  $58.0\pm0.0$ &    - \\
{\textbf{FT}}\textsubscript{RoBERTa} &   84.0 &   79.4 &        71.0 &     80.9 &    92.9 &    64.7 &        94.9 &      58.6 &         91.0 &  $79.7\pm1.9$ &  $-90.0\pm4.7$ \\
{\textbf{P+FT}}\textsubscript{RoBERTa} &   88.2 &   79.6 &        72.5 &     80.8 &    92.7 &    67.0 &        95.2 &      69.8 &         91.1 &  $81.9\pm0.3$ &  $-85.7\pm6.3$ \\
        \bottomrule
        \end{tabular}
    }
    \caption{In-distribution (ID) results for BERT, DeBERTa-v3, and RoBERTa using linear probing (\textbf{LP}), prompting (\textbf{P}), fine-tuning (\textbf{FT}), and prompt-based fine-tuning (\textbf{P+FT}). We report average \textit{Applicability} ($\mu_{F_1}$), \textit{Reliability} ($\mu_{\tau}$), \textit{Stability} ($\sigma_{F_1}, \sigma_{\tau}$). Best OOD performance within one LM are \underline{underlined} and \textbf{bold} highlights best OOD performance across LMs.}
    \label{tab:id-results}
\end{table*}

\subsection{Classification Head Pre-Initialisation}\label{subsec:lpft}
In addition to prompt-based fine-tuning (\textbf{P+FT}), we experimented with pre-initializing the classification head using a linear probe (\textbf{LP+FT}) following \citet{Kumar2022FineTuningCD}.
As reported in \autoref{tab:lpft-results}, we did not find a positive effect of using LP+FT. 

\begin{table*}[t]
\centering
    \setlength{\tabcolsep}{3pt}
    \resizebox{1.00\textwidth}{!}{%
        \begin{tabular}{l|cccc|ccc|cc|cc}
        \toprule
       & \bf arg-qua & \bf arg-sim & \bf arg-cls & \bf evi-cls & \bf review & \bf stance & \bf entail & \bf x-stance & \bf x-review & \bf \textit{Applicability} & \bf \textit{Reliability}\\
     &  \it \small{Top.}  &  \it  \small{Top.} &  \it    \small{Top.}  &  \it  \small{Top.}   &  \it \small{Dom.}   &  \it  \small{Dom.}  &  \it   \small{Dom.}   &   \it    \small{Lang./Top.} &  \it     \small{Lang./Dom.} & $\mu_{F_1}\pm\sigma_{F1}$   &  $\mu_{\tau}\pm\sigma_{\tau}$    \\
        \midrule

{\textbf{FT}}\textsubscript{BERT} &   75.5 &   68.4 &        57.5 &     74.7 &    89.3 &    31.1 &        50.7 &  62.0/63.9 &   77.7/84.4 &  $66.8\pm0.9$ &  $-56.8\pm12.3$  \\
{\textbf{LP+FT}}\textsubscript{BERT} &   75.7 &   66.5 &        57.3 &     74.1 &    89.3 &    34.2 &        50.4 &  60.8/64.1 &   77.1/84.0 &  $66.7\pm1.4$ &  $-56.5\pm14.3$  \\
{\textbf{P+FT}}\textsubscript{BERT} &   76.2 &   66.0 &        59.8 &     75.7 &    89.3 &    28.5 &        48.0 &  59.5/63.6 &   79.6/83.9 &  $66.4\pm1.1$ &  $-61.7\pm12.4$ \\ \midrule

{\textbf{FT}}\textsubscript{DeBERTa-v3} & 78.4 &   75.4 &        64.0 &     77.3 &    93.4 &    29.6 &        55.6 &  69.8/69.3 &   91.3/90.9 &  $72.3\pm1.1$ &  $-72.6\pm13.4$ \\ 
{\textbf{LP+FT}}\textsubscript{DeBERTa-v3} &  78.4 &   75.6 &        63.7 &     76.5 &    93.6 &    30.1 &        54.7 &  69.6/69.1 &   91.1/91.1 &  $72.1\pm1.3$ &  $-70.8\pm11.9$\\ 
{\textbf{P+FT}}\textsubscript{DeBERTa-v3} &    78.5 &   79.1 &        74.6 &     78.6 &    94.2 &    33.0 &        60.2 &  69.7/69.9 &   91.8/91.4 &  $74.6\pm0.9$ &   $-78.4\pm8.4$ \\ \midrule

{\textbf{FT}}\textsubscript{RoBERTa} &   70.9 &   73.0 &        56.9 &     77.5 &    92.2 &    30.0 &        51.3 &  62.2/66.8 &   89.6/90.1 &  $69.1\pm2.5$ &  $-69.7\pm10.4$ \\
{\textbf{LP+FT}}\textsubscript{RoBERTa} &  76.0 &   73.9 &        54.3 &     77.2 &    92.1 &    27.3 &        47.6 &  62.3/67.0 &   89.1/89.2 &  $68.7\pm1.7$ &  $-71.0\pm11.7$ \\
{\textbf{P+FT}}\textsubscript{RoBERTa} &    77.6 &   74.3 &        66.0 &     77.9 &    92.0 &    29.1 &        52.4 &  67.4/67.5 &   89.7/90.0 &  $71.3\pm0.5$ &   $-75.5\pm8.1$  \\
        \bottomrule
        \end{tabular}
    }
    \caption{Comparing vanilla (\textbf{FT}), linear-probing fine-tuning afterward (\textbf{LP+FT}), and prompt-based fine-tuning (\textbf{P+FT}) for BERT, DeBERTa-v3, and RoBERTa. We report average \textit{Applicability} ($\mu_{F_1}$), \textit{Reliability} ($\mu_{\tau}$), \textit{Stability} ($\sigma_{F_1}, \sigma_{\tau}$). }
    \label{tab:lpft-results}
\end{table*}